%% file: example_paper.tex
\documentclass{article}

\usepackage{microtype}
\usepackage{graphicx}
\usepackage{subfigure}
\usepackage{booktabs} % for professional tables
\usepackage{xspace}
\usepackage{hyperref}

\newcommand{\B}{\textit{B}\xspace}
\newcommand{\W}{\textit{W}\xspace}
\newcommand{\gif}{\textit{GIS}\xspace}
\newcommand{\gih}{\textit{GIS-H}\xspace}
\newcommand{\po}{\textit{PipeOffload}\xspace}
\newcommand{\pof}{\textit{PO-F}\xspace}
\newcommand{\poh}{\textit{PO-H}\xspace}
\newcommand{\fb}{\textit{1F1B}\xspace}
\newcommand{\fbi}{\textit{1F1B-I}\xspace}

\usepackage[accepted]{icml2025}

\usepackage{amsmath}
\usepackage{amssymb}
\usepackage{mathtools}
\usepackage{amsthm}

\usepackage{graphicx}
\usepackage{cellspace}
\usepackage[capitalize,noabbrev]{cleveref}

\theoremstyle{plain}

\theoremstyle{definition}

\theoremstyle{remark}

\usepackage[textsize=tiny]{todonotes}

\icmltitlerunning{PipeOffload: Improving Scalability of Pipeline Parallelism with Memory Optimization}

\begin{document}

\twocolumn[
\icmltitle{PipeOffload: Improving Scalability of Pipeline Parallelism\\ with Memory Optimization}



\icmlsetsymbol{equal}{*}

\begin{icmlauthorlist}
\icmlauthor{Xinyi Wan}{equal,sea,nus}
\icmlauthor{Penghui Qi}{equal,sea,nus}
\icmlauthor{Guangxing Huang}{sea}
\icmlauthor{Min Lin}{sea}
\icmlauthor{Jialin Li}{nus}
\end{icmlauthorlist}

\icmlaffiliation{sea}{Sea AI Lab}
\icmlaffiliation{nus}{National University of Singapore}

% \icmlcorrespondingauthor{Xinyi Wan}{wanxy@sea.com}
% \icmlcorrespondingauthor{Penghui Qi}{qiph@sea.com}
\icmlcorrespondingauthor{Min Lin}{linmin@sea.com}
\icmlcorrespondingauthor{Jialin Li}{lijl@comp.nus.edu.sg}

\icmlkeywords{LLM Training, Pipeline Parallelism, Activation Offload}

\vskip 0.3in
]



\printAffiliationsAndNotice{\icmlEqualContribution} % otherwise use the standard text.

\input{paper/0-abstract}

\input{paper/1-intro}

\input{paper/2-stage_level}

\input{paper/3-scheduling}

\input{paper/4-offloading}

\input{paper/5-experiments}

\input{paper/6-relatedwork}

\input{paper/7-conclusion}

\section*{Impact Statement}

This paper presents work whose goal is to advance the field of 
Large Language Model Training Systems. There are many potential societal consequences 
of our work, none which we feel must be specifically highlighted here.


\bibliography{example_paper}
\bibliographystyle{icml2025}

%%%%%%%%%%%%%%%%%%%%%%%%%%%%%%%%%%%%%%%%%%%%%%%%%%%%%%%%%%%%%%%%%%%%%%%%%%%%%%%
%%%%%%%%%%%%%%%%%%%%%%%%%%%%%%%%%%%%%%%%%%%%%%%%%%%%%%%%%%%%%%%%%%%%%%%%%%%%%%%
% APPENDIX
%%%%%%%%%%%%%%%%%%%%%%%%%%%%%%%%%%%%%%%%%%%%%%%%%%%%%%%%%%%%%%%%%%%%%%%%%%%%%%%
%%%%%%%%%%%%%%%%%%%%%%%%%%%%%%%%%%%%%%%%%%%%%%%%%%%%%%%%%%%%%%%%%%%%%%%%%%%%%%%
\newpage
\appendix
\onecolumn

% You can have as much text here as you want. The main body must be at most $8$ pages long.
% For the final version, one more page can be added.
% If you want, you can use an appendix like this one.  

% The $\mathtt{\backslash onecolumn}$ command above can be kept in place if you prefer a one-column appendix, or can be removed if you prefer a two-column appendix.  Apart from this possible change, the style (font size, spacing, margins, page numbering, etc.) should be kept the same as the main body.

\input{paper/appendix}
%%%%%%%%%%%%%%%%%%%%%%%%%%%%%%%%%%%%%%%%%%%%%%%%%%%%%%%%%%%%%%%%%%%%%%%%%%%%%%%
%%%%%%%%%%%%%%%%%%%%%%%%%%%%%%%%%%%%%%%%%%%%%%%%%%%%%%%%%%%%%%%%%%%%%%%%%%%%%%%

\end{document}

%% file: paper/0-abstract.tex
\begin{abstract}

Pipeline parallelism (PP) is widely used for training large language models (LLMs), yet its scalability is often constrained by high activation memory consumption as the number of in-flight microbatches grows with the degree of PP. In this paper, we focus on addressing this challenge by leveraging the under-explored memory offload strategy in PP. With empirical study, we discover that in the majority of standard configurations, at least half, and potentially all, of the activations can be offloaded with negligible overhead. In the cases where full overload is not possible, we introduce a novel selective offload strategy that decreases peak activation memory in a better-than-linear manner. Furthermore, we integrate memory offload with other techniques to jointly consider overall throughput and memory limitation. Our experiments proves that the per-device activation memory effectively reduces with the total number of stages, making PP a stronger alternative than TP, offering up to a 19\% acceleration with even lower memory consumption. The implementation is open-sourced at \href{https://github.com/sail-sg/zero-bubble-pipeline-parallelism}{this url}.
% our experiments showcase significant reductions in activation memory consumption while maintaining throughput levels higher than 1F1B, ultimately enhancing the efficiency and scalability of pipeline parallelism for LLM training.

\end{abstract}

%% file: paper/1-intro.tex
\section{Introduction} \label{sec:intro}

As modern large transformer models \citep{vaswani2017attention} scale towards trillions of parameters, model parallelism becomes essential for distributing model parameters across multiple devices. Compared to ZeRO~\citep{rajbhandari2020zero} and tensor parallelism \citep{shoeybi2019megatron}, pipeline parallel (PP) \citep{huang2019gpipe, harlap2018pipedream} has a lower communication volume and a higher arithmetic intensity. 
However, while PP shards layers across devices to reduce parameter memory requirements, its scalability remains constrained by the activation memory. Increasing the number of PP stages reduces layers per device but necessitates more in-flight microbatches to minimize pipeline bubbles. This trade-off leaves overall activation memory demands unchanged.

In this work, we address this memory limitation of PP by offloading memory to the host. While memory offload is widely adopted in data parallelism (DP) \citep{goyal2017accurate, ren2021zero}, its potential in PP remains largely unexplored. PP is particularly suited for memory offload because the gap between the forward pass and the backward pass creates a natural window for offloading and reloading activation memory without interfering with other computations. This contrasts sharply with activation rematerialization~\citep{chen2016training} which introduces significant recomputation overhead. Memory offload, when properly scheduled and overlapped with other computation, can be a free lunch.

% A key design principle in this paper is to avoid per-microbatch overhead. In other words, \textbf{we allow overhead only during the warmup and cooldown phases, thus keeping the overall impact minimal.}

% \textbf{A surprising finding is that full activation offloading is feasible in many scenarios with negligible overhead. This is crucial for the scalability of PP because activation memory ceases to be a bottleneck when full offloading is implemented.}
% To avoid significant slowdown of the training process, it requires data transfer can be completely overlapped with computations or pipeline bubbles.

% One major concern for activation memory offloading is the relatively low bandwidth between host and device, which might hinder the training efficiency. A common method to mitigate this issue is to overlap the data transfers with the compute \citep{ren2021zero} \citep{yuan2024accelerating}.
% Overlapping data transfers with compute \citep{ren2021zero} \citep{yuan2024accelerating} is a common technique to hide the overhead introduced by offloading methods. However, it requires the time for data transfer to be lower than compute time. Otherwise offloading introduces significant slowdown of the training process. 
Formally, for a single transformer layer, if we define $T_{o}$ as the round-trip time for its activation memory to be moved from device to host (D2H) and then host to device (H2D), $T_{c}$ as the time for its total forward and backward compute, then the ratio between them, denoted as $k=T_{o}/T_{c}$, is an important indicator on the proportion of the activation memory that can be offloaded without introducing significant efficiency drawbacks. Full activation memory offload is possible if $k \leq 1$, as the offload operations can be fully overlapped by compute. Without considering the trivial layers which can be recomputed with negligible overhead (e.g. Dropout, GeLU, LayerNorm), we can estimate the offload cost for the rest of the layers and subsequently estimate $k$. With sequence length ($s$), hidden size ($h$), PCI-E duplex bandwidth ($B_o$) and GPU compute bandwidth ($B_c$), $k$ is computed as: \citep{narayanan2021efficient,korthikanti2023reducing}
% $$T_o=2*\frac{20bsh}{B_o}$$
% $$T_c=\frac{12bsh(6h+s)}{B_c}$$
\begin{equation}\label{formula:k}
k=\frac{T_o}{T_c}=\frac{10}{3(6h+s)}*\frac{B_c}{B_o}    
\end{equation}

\begin{figure*}
    \centering
    \includegraphics[width=0.80\linewidth]{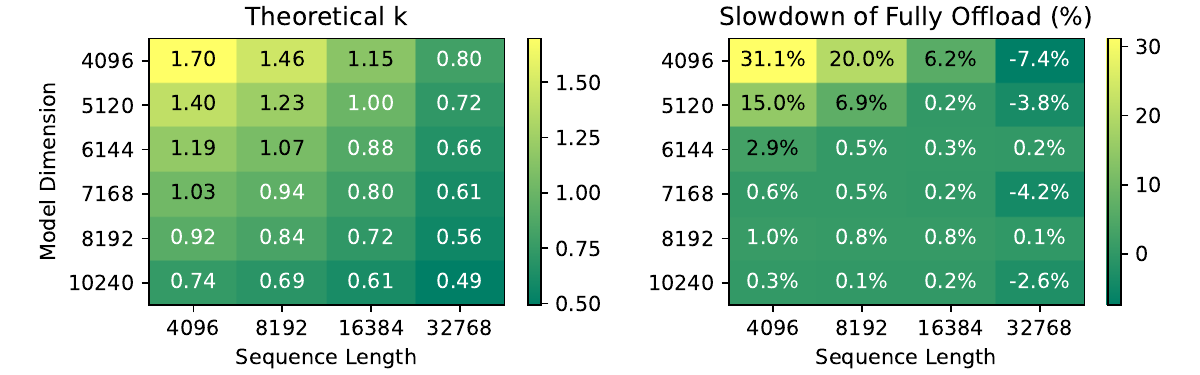}
    \caption{Ablation studies on offload overhead on NVIDIA A100 GPUs. On the left, $k$ values were estimated using Formula \ref{formula:k} with $B_c = 220$ TFLOPS/s and $B_o = 15$ GB/s. On the right, the reduction in throughput due to offload was measured through experiments. The experiments are performed utilizing the fully offloaded 1F1B schedule outlined in Figure \(\ref{app:1f1b_offload}\), involving 8 PP devices and 32 microbatches. The number of transformer layers was chosen to ensure that the baseline, without offload, does not OOM. It is important to note that some values in the second graph are negative, as the baseline experiences frequent CUDA malloc/dealloc operations due to high memory usage.}
    \label{fig:ablation-a100}
\end{figure*}

Note that the value of $k$ decreases as model size or sequence length increases. Figure \ref{fig:ablation-a100} (left) demonstrates that the value of $k$ is surprisingly small under typical hidden dimension and sequence size settings, indicating that a significant portion of activation memory can be offloaded. We observe that $k\leq1$ when the hidden dimension exceeds 8k or the sequence length is greater than 16k; under these conditions, \textbf{all activation memory can be offloaded with negligible overhead}, as shown in Figure \ref{fig:ablation-a100} (right).

When $k>1$, offloading all activation memory would lead to a decrease in throughput. In this scenario, we resort to a partial offload where a subset of the activations are offloaded. Similar to the rematerialization approach where trivial operations are preferred as they impose less computation burden, we perform \textbf{selective offload} that prioritize activations that yield the greatest reduction in peak memory usage.
We introduce a general guideline for selective offload that always prefers activations with longer lifespan, i.e., longer gaps between forward and backward passes. 
% This simple strategy leads to a peak memory reduction that scales at least linearly with the amount of offloading performed.
Intuitively, the longer an activation remains in flight, the more it contributes to peak memory.

\begin{figure}
    \centering
    \includegraphics[width=0.8\linewidth]{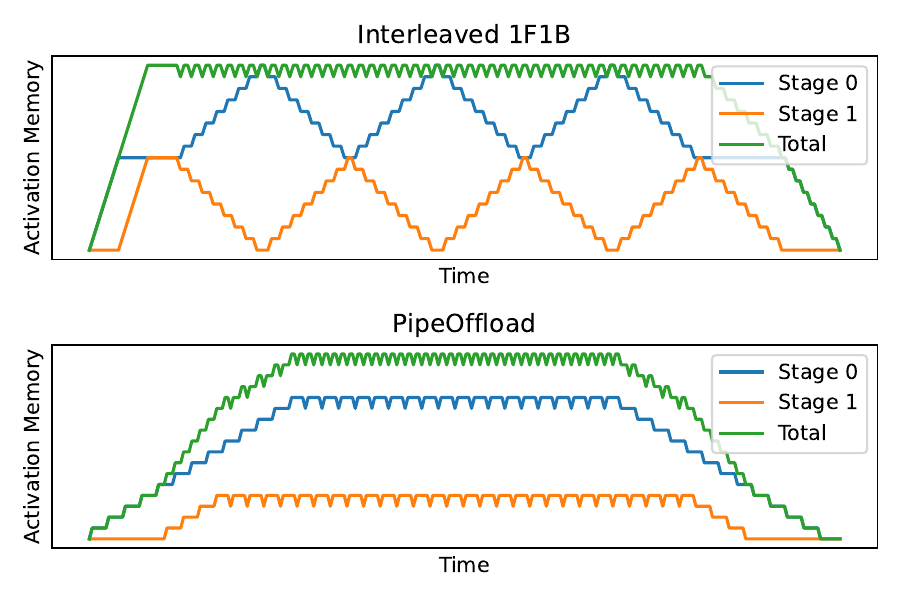}
    \caption{Memory pattern of different schedules. We plot the activation memory of each stage separately and show their contribution to the total activation memory. In Interleaved 1F1B, offloading stage 0 results in only a 50\% reduction in peak activation memory, despite stage 0 having a longer lifespan. Contrastingly, \po at the bottom distributes activation memory uniformly across time, offering better-than-linear memory savings if stage 0 is offloaded.}
    \label{fig:repeat}
\end{figure}

A widely adopted strategy to reduce pipeline bubbles involves placing multiple stages on the same device, as demonstrated in works such as \citep{narayanan2021efficient}, \citep{qi2024pipeline}, and \citep{Liu2023HanayoHW}. Notably, different stages have varying lifespans, potentially resulting in a more efficient, better-than-linear reduction in peak memory usage. The efficiency of selective offload also depends on the memory usage pattern of the pipeline schedule. For instance, when visualizing the memory usage patterns of interleaved 1F1B \cite{narayanan2021efficient} and our \po method in Figure \ref{fig:repeat}, it is apparent that offloading stage 0 in our method yields a 3/4 reduction in peak memory, while in the interleaved 1F1B, at most a 1/2 reduction can be achieved.

When memory offload is applied in practice, we need to further consider its interplay with other factors, especially the trade-off between memory and throughput. We extend the interleaving strategy into a generalized form, offering smooth memory reduction with minimal throughput loss. This approach provides flexibility in optimizing performance based on specific system needs. 

% When memory offload is applied in practice, we need to further consider its interplay with the various other factors. An optimal solution would jointly consider the pipeline bubbles (and hence the throughput), the peak memory, the amount of memory offload and its overlap with computation and other communication operations. In this work, we propose a pipeline schedule that strikes a balance on these factors. 

In the rest of this paper, we detail the selective offload strategy in Section \ref{sec:selective-offloading}, introduce a family of pipeline schedules that trade-off between throughput and peak memory in Section \ref{sec:general_interleaving}, elaborate on implementation details in Section \ref{sec:offloading_pp}, and finally evaluate and compare the methods in Section \ref{sec:exp}.

\begin{figure}
    \centering
    \includegraphics[width=0.98\linewidth]{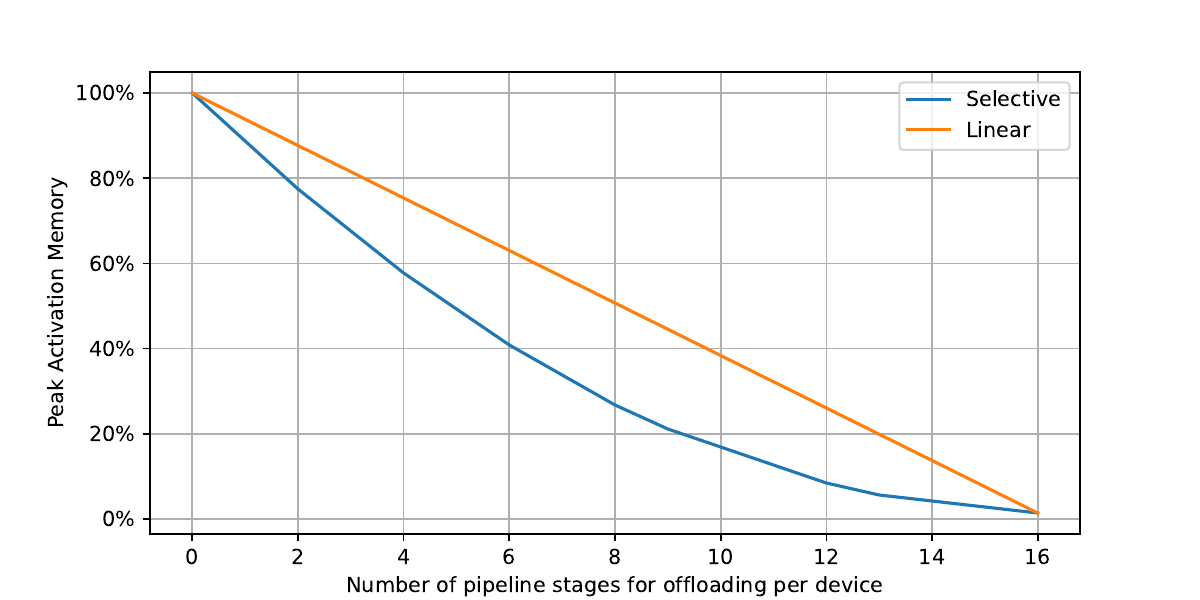}
    \caption{The memory reduction ratio of stage-level offload under 8 PP devices and 16 stages.}
    \label{fig:better_than_linear}
\end{figure}

% To achieve greater memory reduction with the same budget, we employ the uniform repeating strategy from \cite{qi2024pipeline} and introduce a novel stage-level selective offloading/recomputing strategy. This strategy prioritizes offloading or recomputing those early pipeline stages which have longer lifespans. As shown in Figure \ref{fig:better_than_linear}, by offloading only half of the pipeline stages, peak memory can be reduced to about one-quarter in scenarios with 8 PP devices and 16 pipeline stages per device.

% \cite{qi2024pipeline} provides a framework to build a pipeline schedule, in which the peak activation memory is proportional to the sum of the lifespan of each pipeline stage. Intuitively, the longer activations remain in flight, the more memory a PP schedule consumes. 

% It's important to note that the uniform repeating strategy may increase pipeline bubbles. Although the overhead only occurs during the warmup and cooldown phases, reducing these extra pipeline bubbles can still enhance throughput. \textbf{We propose a generalized interleaving strategy to reduce peak memory by shortening the lifespan of interleaving 1F1B, which is conceptually simple and can smoothly reduce peak memory usage by up to half, with fewer pipeline bubbles than uniform repeating.}.

%% file: paper/2-stage_level.tex
\section{Selective Offload} \label{sec:selective-offloading}

When full activation offload is not feasible ($k > 1$) without overhead on the throughput, an efficient selective offload strategy becomes crucial. In this section, we explore considerations for a selective offload strategy when multiple pipeline stages are placed into each device as in interleaved 1F1B \citep{narayanan2021efficient}.

\begin{figure}
    \centering    \includegraphics[width=0.98\linewidth]{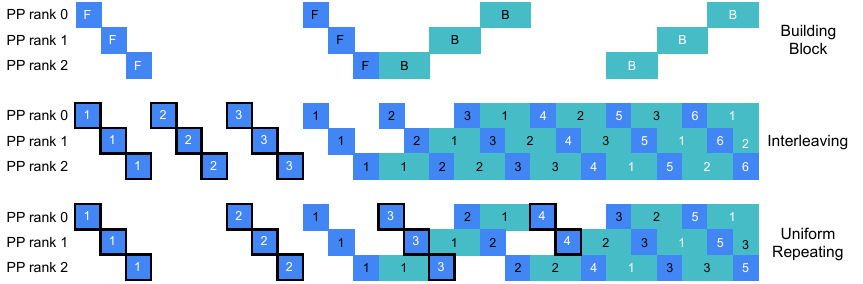}
    \caption{The building block (top) describes the pattern for each microbatch, where \textit{F} represents forward and \textit{B} represents backward. Both the interleaving (middle) and uniform repeating (bottom) strategies adhere to this building block. Although sharing the same peak memory, the contributions of pipeline stages differ in these two strategies. We emphasize the contributions of the first pipeline stages with bold borders.}
    \label{fig:bb_distribution}
\end{figure}

Following \cite{qi2024pipeline}, we decompose a pipeline schedule into repeating a building block, which describes how a single microbatch should be scheduled in the pipeline (as in Figure \ref{fig:bb_distribution}). 
By ensuring each microbatch adheres to this pattern,
the peak memory usage is approximately proportional to the summed lifespan of each pipeline stage. Intuitively, the longer activations remain in flight, the more memory a PP schedule consumes. Based on this insight, preferring offloading pipeline stages with longer lifespan is a natural selective strategy for more memory reduction.

However, lifespan is not the sole factor influencing a stage's contribution to peak memory. The strategy for organizing microbatches within a pipeline schedule also significantly impacts how stages contribute to peak memory.
We evaluate two strategies for building pipeline schedules: the interleaving strategy (as in \cite{narayanan2021efficient}) and the uniform repeating strategy (as in \cite{qi2024pipeline}). The interleaving strategy employs a bi-level repeating pattern: the outer loop interleaves pipeline stages, while the inner loop uniformly repeats a set number of microbatches. In contrast, the uniform repeating strategy consistently schedules the next microbatch after a fixed offset. As shown in Figure \ref{fig:bb_distribution}, although both strategies share the same building block and peak memory, their contributions to peak memory differ. Specifically, offloading or recomputing the first pipeline stage (indicated by white numbers) in rank 0 reduces peak memory by 3 activations in the interleaving strategy, whereas the uniform repeating strategy achieves a reduction of 4 activations. This demonstrates a greater memory reduction with the uniform repeating strategy than with interleaving srategy. 

Figure \ref{fig:repeat} illustrates the contribution of each pipeline stage to peak activation memory across 8 devices. It is evident that in both strategies, stage 0 (with longer lifespan) contributes more than or equal to stage 1 (with shorter lifespan) to peak memory. Notably, the uniform repeating strategy, where each stage's contribution to peak memory is roughly proportional to its lifespan~\citep{qi2024pipeline}, tends to offer greater memory reduction compared to the interleaving strategy under the same budget.

Based on these observations, we propose prioritizing the offloading of pipeline stages with longer lifespans. The uniform repeating strategy should be favored over the interleaving strategy for its superior memory reduction efficiency within the same offload budget. Figure \ref{fig:better_than_linear} presents the theoretical peak memory curve under various offload budgets using the uniform repeating strategy. It demonstrates better-than-linear efficiency in memory reduction. Notably, by offloading only half of the pipeline stages, peak memory can be reduced to approximately one-quarter in scenarios with an 8 PP degree and 16 pipeline stages per device.

% Additionally, the pipeline bubbles are always less than or equal to the bubbles in the building block, because both repeating and squeezing never increase the number of bubbles. So even without reordering, the final schedule is still more efficient than 1F1B.

%% file: paper/3-scheduling.tex
\begin{figure*}
    \centering
    \includegraphics[width=0.98\linewidth]{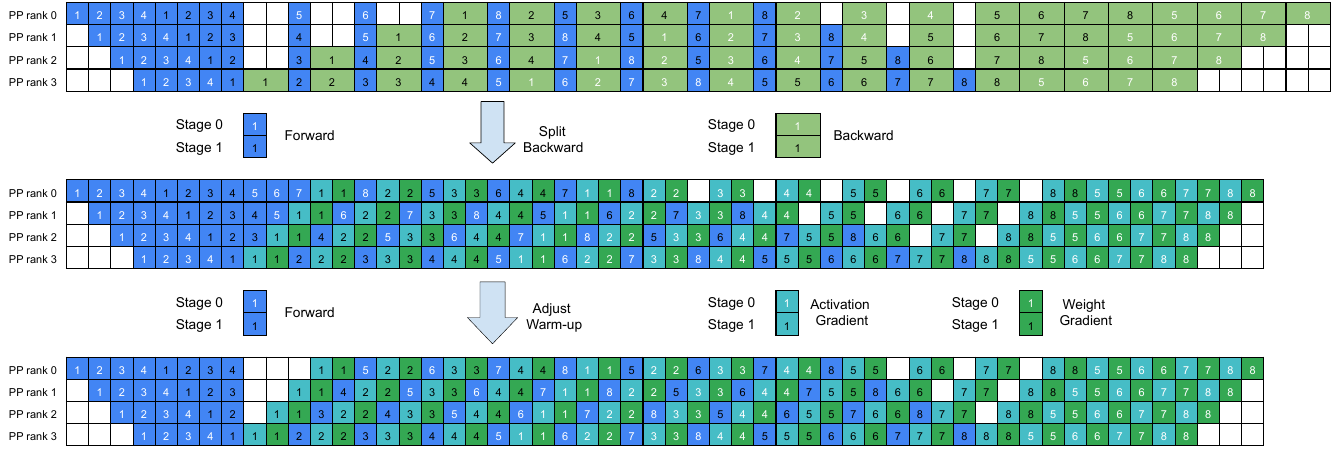}
    \caption{Top: vanilla interleaved 1F1B; Middle: with split backward; Bottom (\gif): after adjusting warmup.}
    \label{fig:adjust_interleave}
\end{figure*}

% \begin{figure*}
%     \centering
%     \includegraphics[width=0.98\linewidth]{figs/group_bw_1.pdf}
%     \caption{Memory efficient interleaved 1F1B with the same pipeline bubbles but lower memory.}
%     \label{fig:group_interleve_1}
% \end{figure*}

\begin{figure*}
    \centering
    \includegraphics[width=0.98\linewidth]{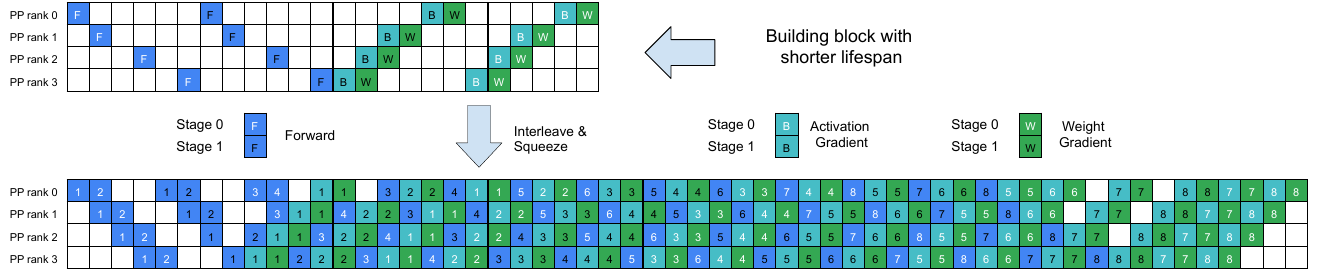}
    \caption{Top: the building block of \gih; Bottom: \gih schedule with less activation memory.}
    \label{fig:interleave_half}
\end{figure*}

% \section{How to Reduce Memory with Minimal Throughput Loss}
\section{Trading off Memory and Throughput} \label{sec:general_interleaving}

In PP, there is a trade-off between activation memory and pipeline bubbles. While the uniform repeating strategy offers superior memory reduction efficiency through offload, it may also lead to an increased number of pipeline bubbles compared to the interleaving strategy, as illustrated in Figure \ref{fig:bb_distribution}. In scenarios where memory pressure is not a primary concern, the interleaving strategy is preferable due to its ability to maintain higher throughput. 
In this section, we extend the interleaving strategy into a generalized form. This approach aims to achieve smooth memory reduction while minimizing throughput loss, providing a flexible solution that can be tailored to specific system requirements.

% In pipeline parallelism, balancing activation memory and pipeline bubbles is crucial. The uniform repeating strategy excels in memory reduction but may introduce more pipeline bubbles than the interleaving strategy (see Figure \ref{fig:bb_distribution}). When memory pressure is low, the interleaving strategy is preferable for higher throughput.
% We extend the interleaving strategy into a generalized form, offering smooth memory reduction with minimal throughput loss. This approach provides flexibility to optimize performance based on specific system needs.

% overall strategy
% Note that the uniform repeating strategy may introduce more pipeline bubbles than the interleaving strategy (Figure \ref{fig:bb_distribution}). However, these bubbles occur only during the warmup and cooldown phases, which are relatively small compared to the per-microbatch overhead.
% Our overall stage-level offloading strategy is to prioritize the generalized interleaving strategy (described in Section \ref{sec:general_interleaving}) due to its smaller overhead. If this approach does not sufficiently reduce activation memory to fit within GPU constraints, we switch to the uniform repeating strategy to achieve greater memory reduction.

\subsection{Free Lunch for Interleaved 1F1B} \label{sec:free_lunch}

% We make several modifications on the vanilla interleaved 1F1B to improve its efficiency, both in pipeline bubbles and activation memory.

\paragraph{Zero-Bubble Strategy}
We use the zero-bubble strategy from \cite{qi2023zero} to reduce pipeline bubbles without any trade-offs. Specifically, the backward pass is split into activation gradient computation (\B) and weight gradient computation (\W). Unlike \textit{ZB-H1} and \textit{ZB-H2} schedules in \cite{qi2023zero}, we don't delay \W passes to further reduce bubbles, as this complicates our strategies. Since our focus is mainly on memory reduction, we simply keep their original schedules with split backward pass (see the middle of Figure \ref{fig:adjust_interleave}).

\paragraph{Adjust Warmup}
We modify the vanilla interleaved 1F1B schedule to make it more memory efficient, as shown in Figure \ref{fig:adjust_interleave}. Specifically, we reduce the number of forward passes in the warmup phase from $d(v-1) + 2(d - i) - 1$ to $d(v-1) + d - i$ for PP rank $i$, where $d$ is the number of devices and $v$ is the number of pipeline stages per device. This change reduces the maximum peak memory (occurring in rank 0) from  $dv + d - 1$ to $dv$, which is a prominent improvement especially for small $v$. Importantly, this modification achieves memory savings without introducing any additional pipeline bubbles.

For convenience, we refer to the schedule after splitting backward and adjusting warmup as \gif.

\subsection{Memory Reduction by Shortening Lifespan} \label{sec:shorten_lifespan}

We generalize the interleaving strategy to accommodate smaller memory constraints with minimal efficiency loss, while maintaining a 1F1B pattern. 
% Following the perspective in \cite{qi2024pipeline}, we decompose a pipeline schedule into repeating a building block, ensuring each microbatch adheres to this pattern. 
Inspired by \cite{qi2024pipeline}, we reduce peak memory usage by shortening the lifespan of the building block.

From Figure \ref{fig:adjust_interleave}, we observe redundant lifespans for each microbatch. For instance, in \gif, although backward passes have tight dependencies within each stage, the waiting time between stages is excessively long. This presents an opportunity to further reduce activation memory. We design new building blocks with shortened lifespans (Figure \ref{fig:interleave_half}) and organize microbatches into a pipeline schedule by repeating these building blocks. Similar to interleaved 1F1B, we use a bi-level repeating pattern: the outer loop interleaves pipeline stages, and the inner loop uniformly repeats a number of microbatches (denoted as $g$). Notably, \gif is a special case where $g$ equals $d$. We can adjust $g$ to control the lifespan, with a minimal value of $\lceil \frac{d}{2} \rceil$ to satisfy dependencies between stages.
Note that reducing the lifespan incurs extra pipeline bubbles during the warmup phase. For any value of $g$ ($\lceil \frac{d}{2} \rceil \leq g \leq d$), the maximum peak activation memory is $g(v-1) + d$ (in rank 0), and the size of extra pipeline bubbles is $(d-g) * (v-1)$. 
% We refer to the schedule in case $g=\lceil \frac{d}{2} \rceil$ as \gih (Figure \ref{fig:interleave_half}).
By selecting the largest value of $g$ that fits within the memory limit, we can achieve an optimal schedule with minimal throughput loss. In the extreme case where $g=\lceil \frac{d}{2} \rceil$ (Figure \ref{fig:interleave_half}), it results in peak memory usage that is about half of the vanilla interleaved 1F1B. We refer to this specific schedule as \gih.

By uniformly repeating the building block (as described in Section \ref{sec:selective-offloading}) of \gih, with minor modifications to prevent collisions \citep{qi2024pipeline}, we present a schedule called \po with similar peak memory. In \po, offloading half of the pipeline stages is referred to as \poh, while offloading all stages is termed \pof. We compare the bubble and activation memory in Table \ref{table:memory_bubble}.
\begin{table}[h]
\tiny
\begin{center}

\caption{Comparing activation memory and bubble rate of different schedules. Additional notations: Activation memory of the entire model ($M$), Time of \textit{F}, \textit{B}, \textit{W} passes of a single stage on a device ($T_F$, $T_B$, $T_W$). Note that for \pof, the activation memory proportionally decreases with $vd$. }

\label{table:memory_bubble}
\begin{tabular}{c|c|c}
\hline
Schedule & Activation & Bubble  \\ 
&Memory& \\
\hline \hline
\fb & $M$ & $v(d-1)(T_F+T_B+T_W)$ \\ \hline
\fbi & $\frac{(v+1)}{v}M$ & $(d-1)(T_F+T_B+T_W)$  \\[3pt] \hline
\gif & $M$ & $(d-1)(T_F+T_B)$ \\ \hline
\gih & $\frac{(v+1)}{2v}M$ & $(d-1)(T_F+T_B)+\frac{(v-1)d}{2}(T_F+T_B-T_W)$ \\[3pt] \hline
\poh & $\approx \frac{(v+2)}{8v}M$ & $ <v(d-1)(T_F+T_B+T_W)$ \\[3pt] \hline
\pof & $O(\frac{M}{vd})$ & $<v(d-1)(T_F+T_B+T_W)$ \\[3pt] 
\hline

\end{tabular}

\end{center}
\end{table}

% It is important to note that our method should be preferred over other memory reduction techniques, such as activation recomputation, because our method only incurs overhead during the warmup phase. In contrast, activation recomputation introduces overhead throughout the entire pipeline. However, our method cannot assist when the activation memory limit is less than IZB-Half. We propose additional methods to handle such cases in Section \ref{sec:offloading_pp} and Section \ref{sec:uniform-repeating}.

%% file: paper/4-offloading.tex
\begin{figure*}[h!]
    \centering
    \includegraphics[width=0.8\linewidth]{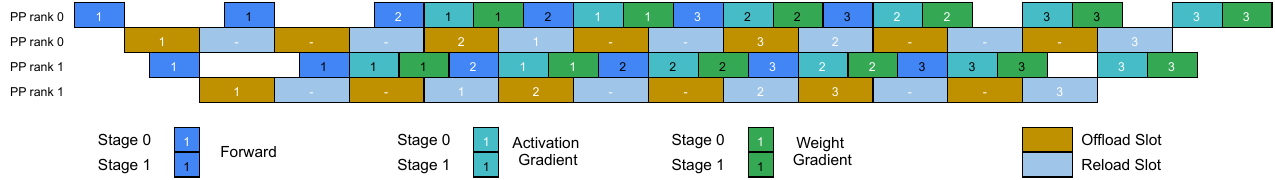}
    \caption{The offload scheduling based on uniform repeating with one pipeline stage offloaded per device and $k=1$. We use a single stream with a one-offload-one-reload pattern, where "-" means empty slot.}
    \label{fig:offload_scheduling}
\end{figure*}

\section{Offload Implementation} \label{sec:offloading_pp}

% We propose to offload activations to host memory to effectively reduce the memory requirements, which can be a powerful augmentation to the generalized interleaving strategy outlined in Section \ref{sec:general_interleaving}. 

In this section, we elaborate the implementation details of our offload strategy, particularly highlighting the differences with the approach described in \cite{yuan2024accelerating}. As we aim to reduce memory usage with minimal overhead, we primarily focus on leveraging the "free lunch" opportunity of offload, which necessitates that the offloading and reloading processes can be fully overlapped with computation, thereby avoiding any additional overhead for the original pipeline.

% We first discuss the offloading bound in Section \ref{sec:offloading_bound}, then illustrate our chunk-level offloading strategy in Section \ref{sec:offload_scheduling}, and finally delve into the key implementation details in Section \ref{sec:implementation}.

\subsection{Improve Offload Efficiency}
To reduce the offload constraint and enable greater offload capabilities, we adopted the subsequent approaches:
a) Employing direct recomputation on activation-heavy but computationally lightweight layers like GeLU to diminish activation memory per layer, leading to a notable 40\% decrease in activation while maintaining throughput efficiency.
b) Guaranteeing a stable and swift PCI-E bandwidth through the utilization of a hardware-topology-aware strategy.
c) Decreasing host-side memory capacity overhead by leveraging continuous buffers.
Furthermore, it is important to highlight that all these methods contribute to reducing $k$, which signifies the number of stages that can be offloaded.
For a more detailed explanation of these techniques, please refer to Appendix \ref{app:implementation}.

\subsection{Offload Scheduling} \label{sec:offload_scheduling}

When integrating offloading and reloading into a pipeline, careful scheduling alongside computation passes is essential. \cite{yuan2024accelerating} employs a fixed scheduling strategy, initiating offloading immediately after the forward pass and reloading at the start of the last backward pass. Offloading and reloading are placed into separate streams to enable overlap. However, in practice we find that separate streams can lead to significant latency fluctuations (see Appendix \ref{sec:memory_transfer_ablation_study}), which can result in notable overhead due to computation passes waiting for offloading or reloading to complete. 
In contrast, we use a single stream for both offloading and reloading. By sharing the stream, we stabilize the latency of offload and reload passes, simplifying the scheduling and enhancing system robustness and performance.

When scheduling the offload and reload passes based on uniform repeating strategy, as shown in Figure \ref{fig:offload_scheduling}, we maintain pipeline bubbles after repeating the building block. For a given model and its training configurations, we first calculate $k$ as defined in Formula \ref{formula:k}. Then, we allocate separate slots in the stream using a one-offload-one-reload pattern.
Given the pipeline stages to offload, we process them one by one from left to right, finding the earliest available offload slot after the forward pass and placing the offload there. For reloading, we move from right to left, identifying the latest available reload slot before the backward pass and placing the reload there. Although some slots may remain unoccupied and the schedule is not squeezed, it can be optimized automatically when running on devices.
For A100 GPUs, where PCI-E is often shared by two adjacent devices, we stagger the offload streams and insert synchronization events across the two devices to avoid the same operation occurs simultaneously (see Appendix \ref{sec:memory_transfer_ablation_study} for more details).

\subsection{$k$ on Other Hardware Platforms} \label{sec:offloading_bound}
% Although the bandwidth between GPU and host is much smaller than the computation speed of GPU, as the model size and context window grow larger, the computation workload typically increases much faster than activation memory. For instance, training GPT-3~\citep{brown2020language} style LLM models in A100 GPU, the offload overhead can be completely overlapped with the computation when the hidden dimension exceeds 8192. TODO(@xinyi)
In Figure \ref{fig:ablation-a100} we showed that $k$ is relatively small on A100 GPUs. While the value of $k$ is reliant on the hardware architecture, we anticipate that H100 will exhibit a similar $k$ to A100, given that H100 boasts 2x PCI-E bandwidth following an upgrade from PCI-E v4 to v5, along with a 3x increase in compute bandwidth. Furthermore, the model flops utilization (MFU) typically reported by the community is lower for H100s (43\% in Llama 3 \citep{dubey2024llama}) in contrast to A100s (approximately 60\% in our evaluations), thereby narrowing the bounds even further.

\subsection{Caveats of Offload}
Though a carefully implemented offload strategy brings only negligible overhead to the compute throughput, there're some moderate issues. 
Firstly, the host memory capacity is another notable bound for the offload. However, host memory is usually several times larger than the total GPU memory installed on the host and is usually extensible with much lower cost compared to GPU.
Secondly, achieving a "free lunch" offload scenario becomes unattainable if the time interval between matching forward and backward passes is shorter than the round-trip time of offload. A low time interval implies a short lifespan, leading to a negligible contribution to peak activation memory. In practice, skipping offloading for these passes has been observed not to impact peak activation memory.
Lastly, the PCI-E traffic for data movement between host and device may interfere with cross-node P2P communication protocols such as Infiniband or RoCE, potentially slowing down communication if not scheduled prudently. Notably, P2P communication does not significantly impact the utilizable PCI-E bandwidh due to two reasons:
a) The volume of P2P communication is substantially lower than offload. For each transformer layer, the total activation memory is ten times greater compared to the layer output.
b) Most P2P communication occurs within a node on NVLINK, bypassing PCI-E, which mitigates the impact on offload constraints.

%If we define $I_{ij}$ as the aforementioned interval on PP rank $i<d$ and chunk $j<v$, a necessary condition for a chunk suffer from the interval bound is $I_{ij}<=T_o=k(T_F+T_B+T_W)$. Moreover, for the generalized interleave 1F1B defined in \ref{sec:general_interleaving} we have $I_{ij} >= (vd-j*d-i-1)*(T_F+T_B)$ because of the dependencies between forward and backward passes. Combining the previous two inequality function we have $k(T_F+T_B+T_W)>=(vd-j*d-i-1)*(T_F+T_B)$ 

%% file: paper/5-experiments.tex
\section{Experiments} \label{sec:exp}
% We organize our experiments to prove three conclusions: a) In Section \ref{sec:same_setting}, we compare memory and throughput of different settings; b) In Section \ref{sec:mem_scaling}, we demonstrate the improved scalability for our methods; c) In Section \ref{sec:grid_search}, we prove our methods can bring acceleration in distributed training.

% \subsection{Environments}

We evaluate our methods on GPT-3-like models based on Megatron-LM \citep{narayanan2021efficient}. In most cases, one transformer layer is removed from both the first and last pipeline stages to address imbalances caused by vocabulary layers, similar to Llama 3 \citep{dubey2024llama} and Deepseek v3 \citep{liu2024deepseek}. The models used are listed in Table \ref{table:experiment_model}. Our primary metrics are throughput, measured as model flops utilization (MFU), and activation memory, defined as the difference between peak and iteration-start memory. The reported activation memory refers to the maximum peak activation memory observed across all devices.

% We evaluate our methods on a series of GPT-3-like models detailed in Table \ref{table:experiment_model} with an implementation based on Megatron-LM project \citep{narayanan2021efficient}. In most cases, we deduct one transformer layer from both the first and last pipeline stages to mitigate imbalances caused by vocabulary layers, similar to the settings in Llama 3 \citep{dubey2024llama} and Deepseek v3 \citep{liu2024deepseek}. The models in our experiments are listed in Table \ref{table:experiment_model}. Our primary metrics include throughput, quantified as model flops utilization (MFU), and activation memory, calculated as the difference between peak memory and iteration-start memory. The activation memory presented in following sections refers to the maximum peak activation memory across all devices.

\begin{table}[h]
\begin{center}

\caption{A list of models used in experiments. For all models we turn on GQA \cite{ainslie2023gqa} with number of query group set to 8.}

\label{table:experiment_model}
\begin{tabular}{c|c|c|c|c|c}
\hline
Model & Layers & Attention & Hidden & Batch & GPUs  \\ 
&&Heads&Size&Size& \\
\hline \hline
5.8B & 32& 32&4096&32 & 2-32 \\
10.5B   & 38 & 40  & 5120 & 64 & 8  \\ 
18.1B   & 46 & 48  & 6144  & 128 & 16 \\ 
42.9B   & 62 & 64  & 8192 & 256 & 32 \\ 
66.6B   & 62 & 80  & 10240 & 256 & 32 \\ 
83.8B   & 78 & 80  & 10240 & 256 & 32 \\ 
\hline

\end{tabular}

\end{center}
\end{table}

Our experiments run on up to 32 NVIDIA A100 80G GPUs on 4 nodes interconnected by RoCE RDMA network. As mentioned in Section \ref{sec:offloading_bound}, though the $k$ is hardware-dependent, we focus on A100 because it is similar for other modern hardware, such as H100.

The pipeline schedules we evaluate include: 
a) \fb \citep{harlap2018pipedream} and vanilla Interleaved 1F1B (referred to as \fbi) \citep{narayanan2021efficient} implemented in Megatron-LM; 
b) Our generalized interleaved schedule, \gif and \gih, detailed in Section \ref{sec:general_interleaving}. 
c) Our better-than-linear offload techniques outlined in previous Sections. We concentrate on two key configurations, \poh and \pof, where either half($\lceil\frac{v}{2}\rceil$) or full($v$) stages are selectively offloaded. Notice that we skip the \pof settings if the corresponding $k>1$.
For all schedules except \fb, we set the number of stages on each device to the maximum possible value so that each stage has at most 1 transformer layer, unless explicitly specified.

\subsection{Activation Memory Reduction with Similar Throughput} \label{sec:same_setting}
% In Figure \ref{fig:same_settings}, we compare the activation memory and throughput across various methods. Compared to \fbi, our \gih moderately reduces activation memory by half, while \poh significantly slashes activation memory to at most \(\frac{1}{6}\). \pof drives activation memory even lower, addressing scenarios where other methods encounter out-of-memory issues. 

% Regarding throughput, generally all methods have a good throughput because all of them contain pipeline bubbles only during warmup and cooldown phases. Out \gif method demonstrates both higher throughput and less activation memory than \fbi. While \poh and \pof exhibit a slightly reduced throughput compared to \gif and \gih, attributed to more extra bubbles, in general they still achieves better throughput than \fb (except for the case where $k$ exactly equals to 1). For detailed numbers, and explanations for missing data points, please refer to Appendix \ref{app:same_settings_detail}.

\begin{figure*}[h]
    \centering
    \includegraphics[width=0.98\linewidth]{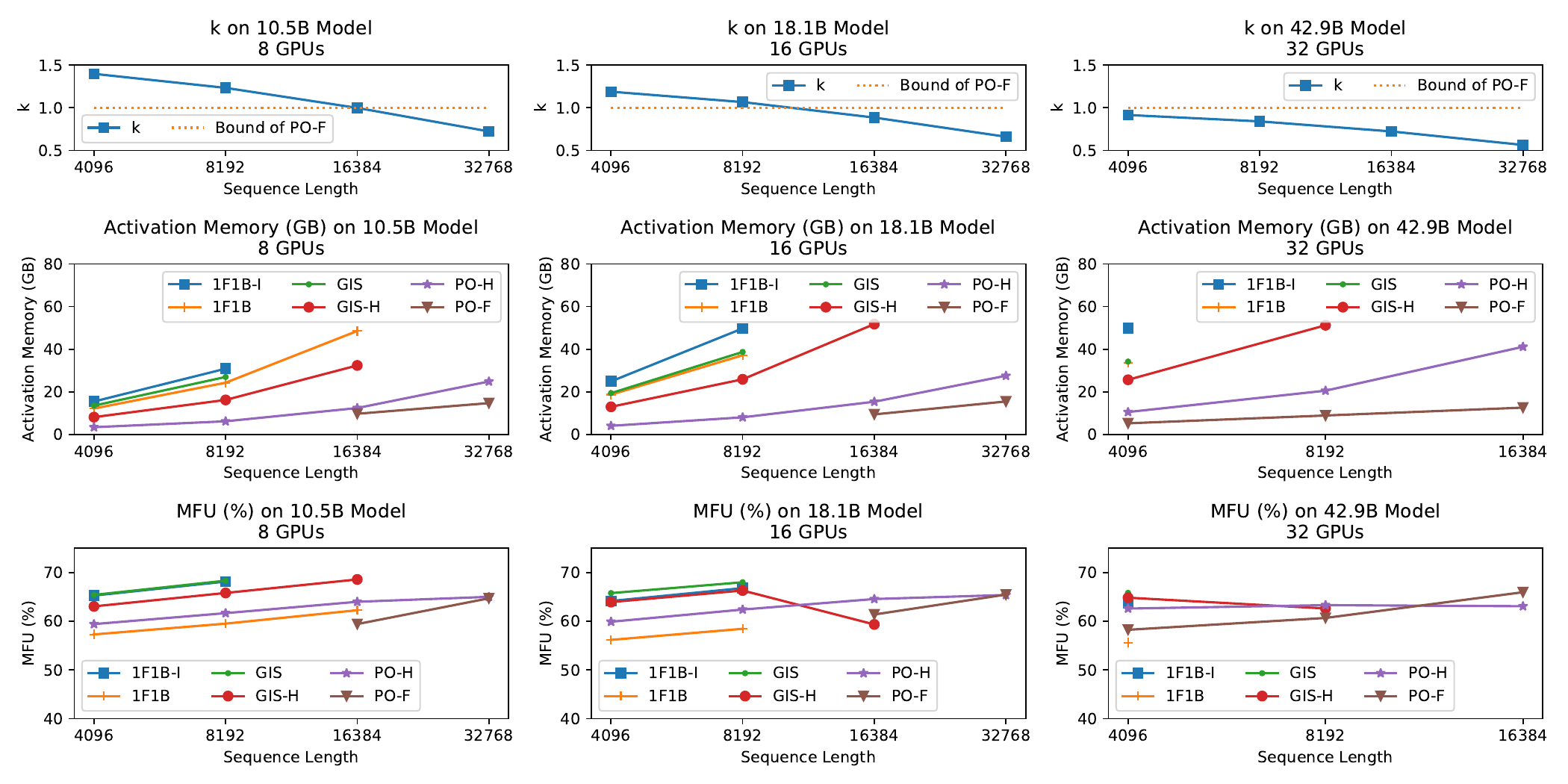}
    \caption{Memory and Throughput Comparison of Different Methods. For detailed numbers, please refer to Figure  \ref{fig:same_settings_detail} in appendix.}
    \label{fig:same_settings}
\end{figure*}

In Figure \ref{fig:same_settings}, we present a comparative analysis of activation memory and throughput across various methods. Compared to \fbi, our \gih method effectively halves the activation memory, while \poh achieves a more substantial reduction, decreasing it to \(\frac{1}{6}\) at most. \pof further minimizes activation memory, providing a solution for scenarios where other methods might encounter out-of-memory issues.

In terms of throughput, all methods generally perform well, with pipeline bubbles only occurring during the warmup and cooldown phases. Our \gif method outperforms \fbi by offering both higher throughput and reduced activation memory. Although \poh and \pof show slightly reduced throughput compared to \gif and \gih, due to additional pipeline bubbles, they still surpass \fb in throughput (except for cases which runs \pof with $k$ exactly equals to 1). For detailed numerical results and explanations of any missing data points, please refer to Figure  \ref{fig:same_settings_detail} in appendix.

\subsection{Better-Than-Linear Selective Offload} \label{sec:super_linear}
\begin{figure*}
    \centering
    \includegraphics[width=0.9\linewidth]{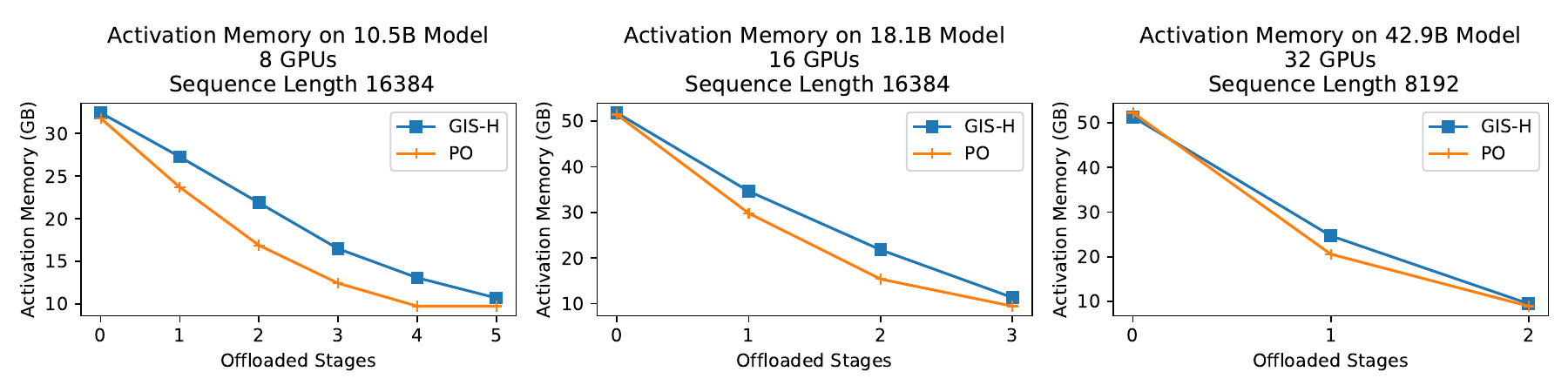}
    \caption{Better-than-linear selective offload. We gradually increase the number stages to offload on two schedules: \gih defined in Section \ref{sec:general_interleaving} and Pipeline Offload (PO), the uniformly repeated schedule introduced in Section \ref{sec:selective-offloading}}
    \label{fig:super_linear}. 
\end{figure*}

In Figure \ref{fig:super_linear}, we illustrate the impact of memory savings when implementing stage-level offload across various schedules. The results indicate that our pipeline offload (PO) schedules with uniform repeating demonstrate memory savings that surpass linear scaling, unlike \gih, aligning with the analysis in Section \ref{sec:selective-offloading}. It's important to note that across all schedules, there exists a consistent overhead caused by temporary activation memory when offloading all stages.

\subsection{Activation Memory Scaling Study} \label{sec:mem_scaling}
\begin{figure}[h!]
    \centering
    \includegraphics[width=0.98\linewidth]{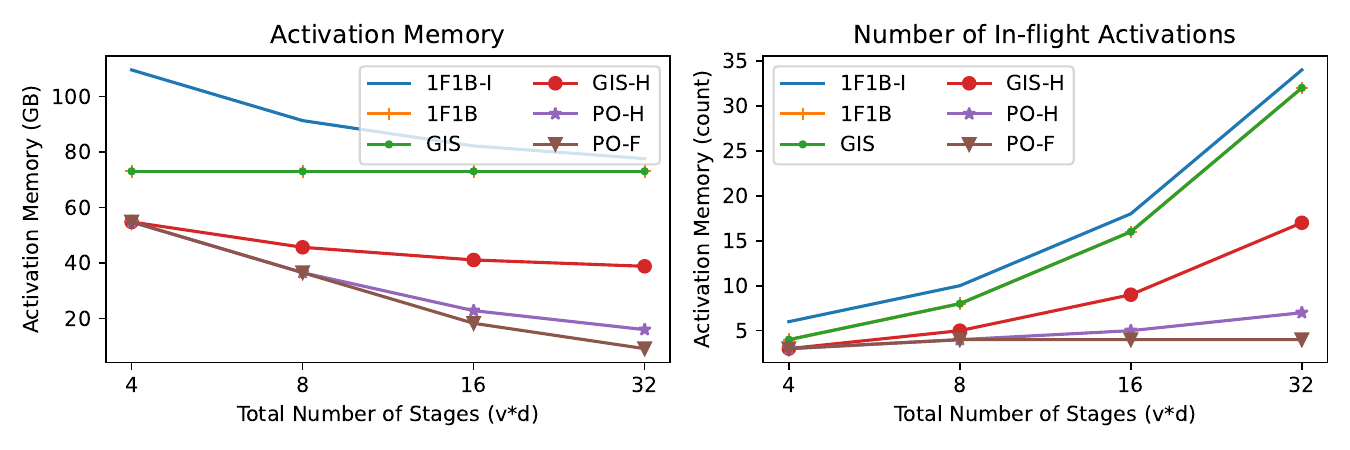}
    \caption{Per-device activation memory when training a 5.8B model using different total number of stages. The left figure shows the activation memory in GB while the right figure shows the number of in-flight microbatches (different stages of the same microbatch are counted multiple times). If there're multiple settings for the same $v*d$, the setting with minimum activation memory is reported. The amount of activation memory is estimated by running the scheduler and count the in-flight activation memory on GPU. Detailed data on this experiment is shown in Figure \ref{fig:scaling_detail} in appendix.}
    \label{fig:scaling_best}
\end{figure}

We delve into the strong scaling of activation memory by analyzing per-device activation memory using a fixed 5.8B model across various total numbers of stages \(v \times d\). The results depicted in Figure \ref{fig:scaling_best} reveal that the per-device activation memory (left figure) of \poh and \pof exhibits superior scaling compared to other methodologies, primarily due to the reduction in the number of in-flight activations (as indicated in the right figure). Notably, the number of in-flight activations remains constant for \pof, matching the analysis in Table \ref{table:memory_bubble}. This observation implies that in the most common scenarios where each stage has only 1 transformer layer, each device essentially maintains an activation memory equivalent to a small constant number (approximately 4 in our experiments) of transformer layers, irrespective of the total number of layers and pipeline devices.

% Specifically, if we consider the activation memory of the entire model as $M$ and the number of layers as $l$, the per-device activation memory for \poh is approximately $\frac{M}{4}$. On the other hand, assuming the number of stages on each device is always the maximum possible value, \pof exhibits an activation memory of around $4*\frac{M}{l}$, indicating that each device essentially has the activation memory equivalent to 4 transformer layers, regardless of the total number of layers and pipeline devices.

\subsection{Comparing With Tensor Parallelism} \label{sec:grid_search}

% In the grid-search experiment we also evaluate with tensor parallel (TP) \citep{shoeybi2019megatron} and sequence parallel (SP) \citep{korthikanti2023reducing}. Unless explicitly specified in later sections, using TP implies using TP+SP.

\begin{figure}
    \centering
    \includegraphics[width=1\linewidth]{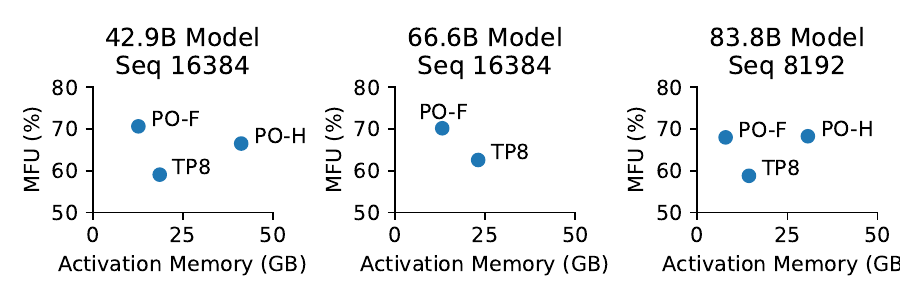}
    \caption{Comparison of pure PP using our methods with hybrid parallelism using PP+TP. \pof and \poh runs as 32-way pure PP while \fbi runs with TP8xPP4.}
    \label{fig:grid_search}. 
\end{figure}

The high activation memory volume is one of the biggest concern for scaling PP to more devices. Commonly, standard settings such as Llama3 \cite{dubey2024llama} often employ a maximum TP degree, typically set at 8, to reduce the activation memory per device. By leveraging our methods to save activation memory on PP, we now compare the performance of using pure PP, which was previously unattainable without our techniques, with interleaved 1F1B combined with 8 TP (together with sequence parallelism in \citep{korthikanti2023reducing}). The results depicted in Figure \ref{fig:grid_search} showcase a notable 12\%-19\% acceleration in training, attributed to the elimination of TP, which typically incurs significant communication overhead. We also highlight that \pof method not only exhibits higher throughput but also consumes less activation memory compared to \fbi combined with the maximum TP degree. This finding suggests that in scenarios where \pof is applicable ($k\leq1$), pure PP should be preferred.

% Besides replacing TP, the significant reduction of activation memory can also be avoid activation recomputation \citep{chen2016training} which incurs at around 30\% slowdown as shown in the experiments of \cite{kim2023bpipe} \cite{yuan2024accelerating} and \cite{qi2024pipeline}. These experiments are well-documented in the literature thus omitted from our paper.

\subsection{Comparing With Other Pipeline Parallel Methods} \label{sec:frontier}

\begin{figure}[H]
    \centering
    \includegraphics[width=1\linewidth]{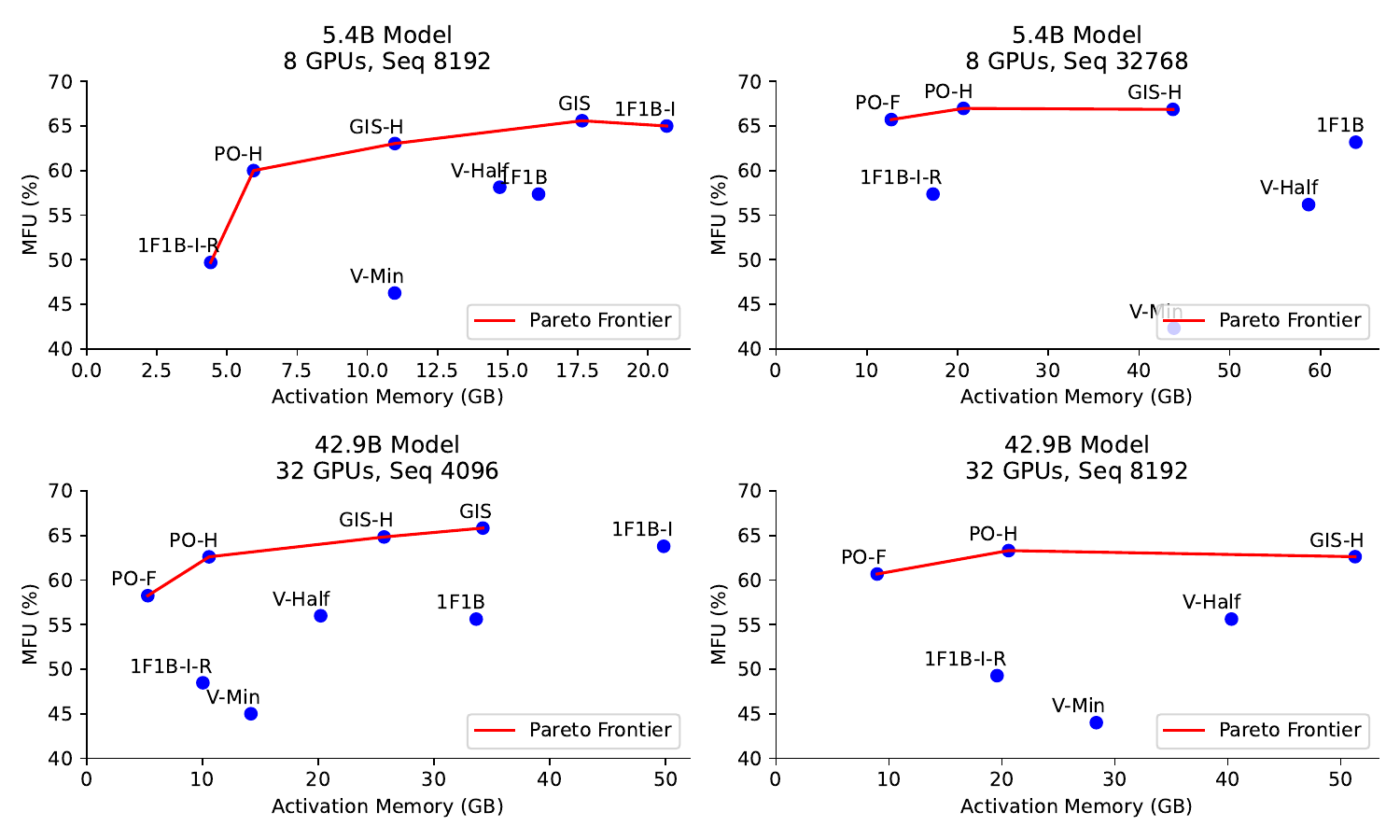}
    \caption{Pareto frontier compared with other methods.}
    \label{fig:frontier}. 
\end{figure}

We compare our approach with existing pipeline parallelism (PP) methods, particularly those aimed at reducing activation memory, including V-Min and V-Half from \citet{qi2024pipeline}, as well as activation re-materialization combined with 1F1B-I (shown as 1F1B-I-R). As shown in Figure \ref{fig:frontier}, our methods achieve a more favorable Pareto frontier in the trade-off between memory usage and throughput. Through detailed analysis, we attribute the suboptimal performance of the V-Shaped schedule proposed in \citet{qi2024pipeline} to significant disparities among $T_F$, $T_B$, and $T_W$ at large sequence lengths. These disparities violate the core assumption in \citet{qi2024pipeline} that these timings are approximately equal, leading to degraded efficiency.

\subsection{Convergence Experiment} \label{sec:frontier}

\begin{figure}
    \centering
    \includegraphics[width=1\linewidth]{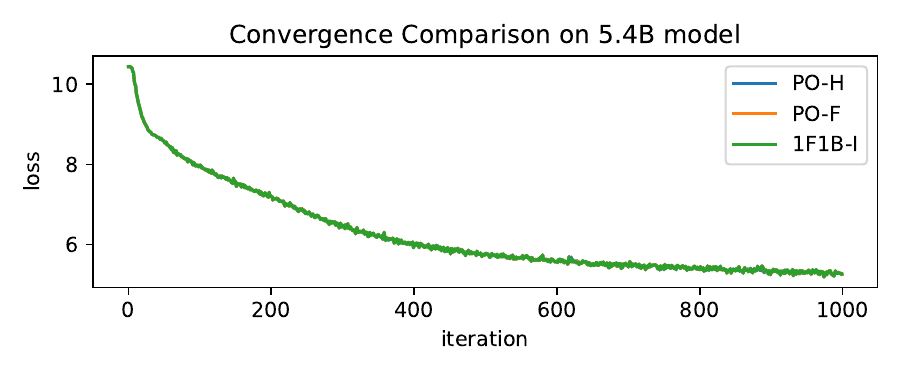}
    \caption{Loss curves proves algorithmic correctness of our methods.}
    \label{fig:convergence}. 
\end{figure}

We compare the loss curves of our \poh and \pof implementations with 1F1B-I to validate correctness on a 5.4B model using 8 GPUs. As shown in Figure~\ref{fig:convergence}, the results exhibit nearly identical loss trajectories, providing strong evidence for convergence correctness. Notably, our methods are designed as exact algorithms with no compromises in accuracy.

% \begin{table*}[h]
% \caption{Comparison between Vanilla interleaved 1F1B and IZB-1 and IZB-Half.}
% \begin{center}
% \begin{tabular}{c|c|c|c|c|c|c}
% \hline
% Model& GPU & $v$ & Seq & Method & Throughput  & Activation Memory (GB)  \\ \hline \hline

% \multirow{9}{*}{5.4B} & \multirow{9}{*}{8} & \multirow{9}{*}{4}
%     & \multirow{3}{*}{4096}
%       &  Vanilla  & 1.218 & 8.86 \\ 
% & & & &  IZB-1    & 1.223 & 7.13 \\ 
% & & & &  IZB-Half & 1.179 & 3.35 \\ \cline{4-7}
% & & & \multirow{3}{*}{8192}
%       &  Vanilla  & 0.560 & 20.3 \\ 
% & & & &  IZB-1    & 0.563 & 16.9 \\ 
% & & & &  IZB-Half & 0.541 & 9.3 \\ \cline{4-7}
% & & & \multirow{3}{*}{16384}
%       &  Vanilla  & 0.234 & 43.0 \\ 
% & & & &  IZB-1    & 0.235 & 36.4 \\ 
% & & & &  IZB-Half & 0.224 & 21.2 \\ \hline

% \multirow{10}{*}{18.1B} & \multirow{10}{*}{16} & \multirow{10}{*}{3}
%     & \multirow{3}{*}{4096}
%       &  Vanilla  & 0.402 & 23.3 \\ 
% & & & &  IZB-1    & 0.410 & 17.0 \\ 
% & & & &  IZB-Half & 0.399 & 9.82 \\ \cline{4-7}
% & & & \multirow{3}{*}{8192}
%       &  Vanilla  & 0.171 & 51.1 \\ 
% & & & &  IZB-1    & 0.192 & 38.6 \\ 
% & & & &  IZB-Half & 0.186 & 24.0 \\ \cline{4-7}
% & & & \multirow{4}{*}{16384}
%       &  Vanilla  & - & OOM \\ 
% & & & &  IZB-1    & - & OOM \\ 
% & & & &  IZB-Half & - & OOM \\ 
% & & & &  Offload  & 0.080 & 34.1 \\ \hline

% \end{tabular}
% \end{center}
% \label{table:profiled_times}
% \end{table*}

%% file: paper/6-relatedwork.tex
\section{Related Work} \label{sec:related_work}

% \# checkpointing, sequence parallelism/selective checkpointing, flash-attention, ZeRO
% Additionally, there is a lot of work which can be used to reduce the overall memory. ZeRO \citep{rajbhandari2020zero} is usually used together with DP, reducing the memory redundancy by sharding parameters across devices. FlashAttention \citep{dao2022flashattention, dao2023flashattention} designs a faster and more memory efficient exact attention algorithm by leveraging different levels of fast and slow memory. Activation recomputation \citep{chen2016training} is a powerful tool to trade computation for memory, however, it may not be desired in many scenarios because it sacrifices the throughput a lot. \cite{korthikanti2023reducing} proposes sequence parallelism to further reduce the activation memory redundancy in Transformer \citep{vaswani2017attention} together with TP. Despite various work, memory shortage is still a significant problem in the training of large models. Our method is orthogonal to these existing methods, and we believe it can further mitigate the memory shortage issue.

\paragraph{Pipeline Parallelism}

Various pipeline schedules have been developed to reduce activation memory usage in PP. An early notable work is 1F1B \citep{fan2021dapple, harlap2018pipedream}, which uses a one-forward-one-backward pattern to mitigate the high memory usage of GPipe \citep{huang2019gpipe}. BPipe \citep{kim2023bpipe} was later introduced to address the memory imbalance issue of 1F1B by transferring activations across devices.
Leveraging the auto-regressive property of causal transformers, token-level pipeline schedules have been proposed by \cite{li2021terapipe, sun2024seq1f1b}, showing promising memory reduction results, especially for long-context training. Vocabulary parallelism was recently proposed by \cite{yeung2024balancing} to address the memory imbalance caused by vocabulary layers, alleviating the memory bottleneck of PP.
\cite{qi2024pipeline} introduced a general framework showing that peak memory can be directly controlled by lifespan of the building block. Based on this insight, they proposed a memory-balanced V-Shape schedule, reducing peak activation to at most 1/3 compared to 1F1B.

\paragraph{Activation Rematerialization and Offload}

Activation rematerialization was first proposed by \cite{chen2016training} which trades computation for memory. To alleviate its overhead, selective strategies have been developed \cite{korthikanti2023reducing, yuan2024accelerating}, focusing on recomputing operations with high memory-to-computation ratio. 

Offload techniques have also been explored to address memory constraints in training LLMs, but prior work \citep{ren2021zero, rajbhandari2021zero} often focuses on model states, leading to poor overlap between data transfer and computation and resulting in high overhead. A recent work \cite{yuan2024accelerating} on offloading activation memory in PP is the most related work to ours. However they draw an opposite conclusion than ours, that activation offload causes significant overhead and should be avoided if possible. They emphasize memory reduction, allowing offload to delay computation, which often results in a brittle schedule introducing significant pipeline overhead. In contrast, we focus on improving the memory reduction efficiency and minimizing overhead by fully overlapping offload with computation. We deliver a different insight that offload can be a free lunch in PP, and full activation is often feasible to make PP scalable.

%In contrast, our approach minimizes overhead by fully overlapping offloading with computation. We introduce a novel stage-level selective offloading strategy that enhances memory reduction efficiency. To the best of our knowledge, our work is the first selective strategy for offloading, achieving greater memory reduction with the same budget.
%Experimental results show that our strategy can significantly improve the scalability of pipeline parallelism, with minimal overhead.

%% file: paper/7-conclusion.tex
\section{Conclusion}
In this work, we present \po, a novel pipeline schedule that incorporates multiple innovative techniques to significantly decrease the activation memory requirements of PP. Through evaluation, we demonstrate that \poh can reduce activation memory to less than a quarter of that required by interleaved 1F1B schedules while maintaining similar throughput across a wide range of real-world models. \pof proves particularly impactful for larger models or extended sequence training tasks, where activation memory can be further reduced to that of a small constant number of transformer layers. These methods greatly improved the scalability of PP and PP becomes feasible in cases they were not possible previously. We demonstrate that the enhanced PP can serve as a compelling alternative to other distributed training methods, such as tensor parallelism.

%% file: paper/appendix.tex
\section{1F1B schedule with Fully Activation Offload} \label{app:1f1b_offload}
\begin{figure*}[h!]
    \centering
    \includegraphics[width=0.98\linewidth]{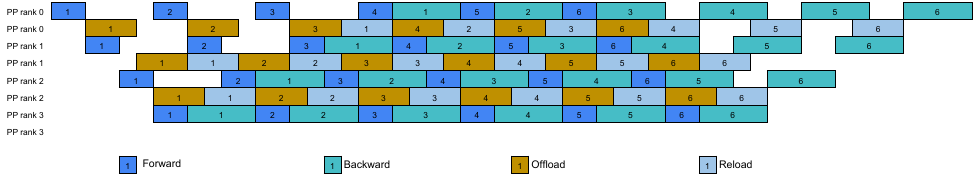}
    \caption{Schedule of applying fully activation offload to 1F1B. Note that in this schedule we also use the topology-aware offload which synchronizes offloading and reloading cross two adjacent devices.}
    \label{fig:1f1b_offload}
\end{figure*}

\section{More Details on Memory and Throughput Comparison between Different Methods} \label{app:same_settings_detail}
\begin{figure*}
    \centering
    \includegraphics[width=0.98\linewidth]{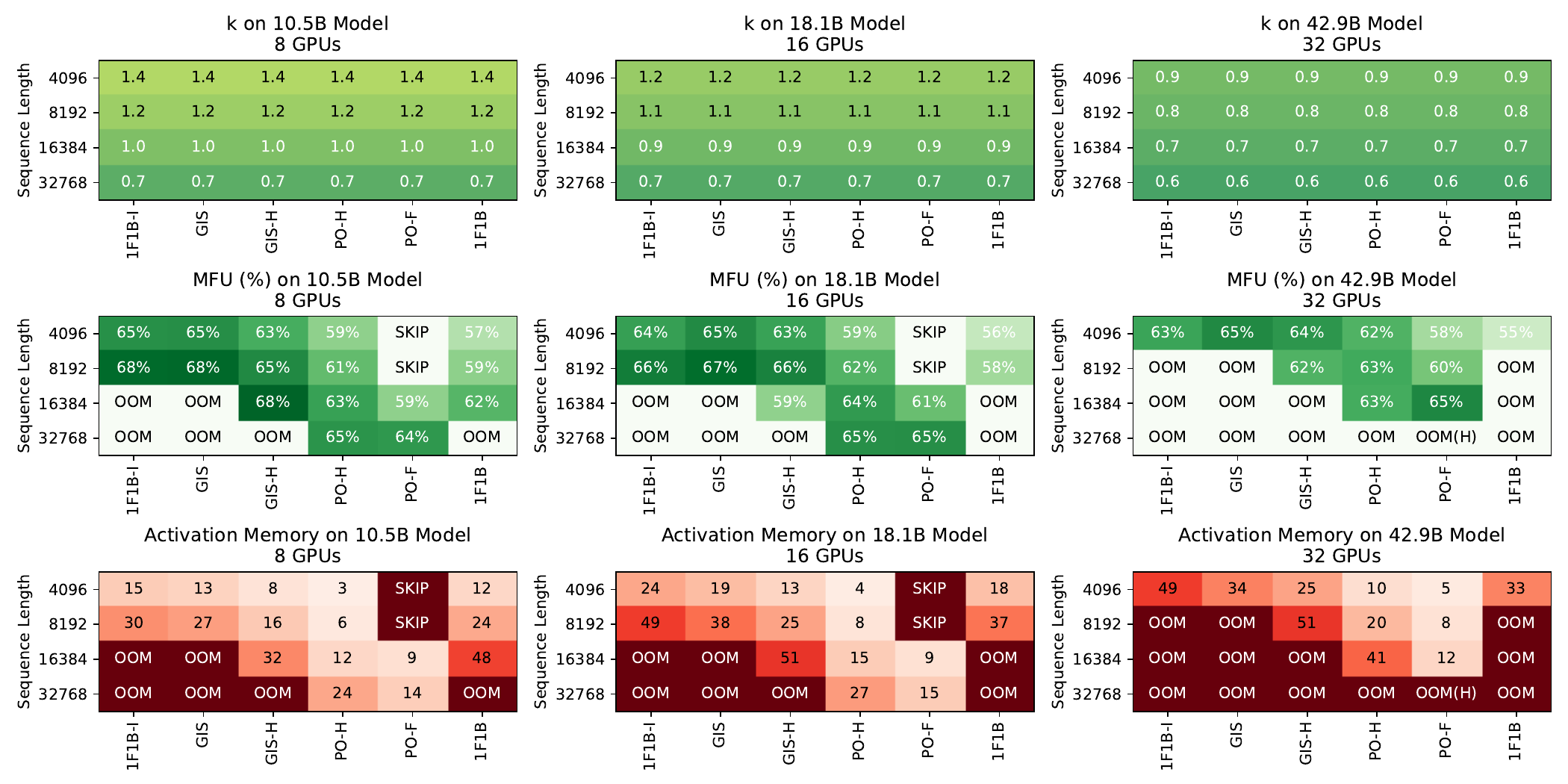}
    \caption{Detailed Data of Throughput and Memory Comparison of Different Methods. SKIP indicates the \pof method is skipped because $k>1$. While OOM indicates a GPU out-of-memory error, OOM(H) represents an OOM on host, which only happens on the largest model with largest sequence length.}
    \label{fig:same_settings_detail}
\end{figure*}

In Figure \ref{fig:same_settings_detail}, we show more details and numbers of the experiment introduced in \ref{sec:same_setting}

\section{Implementation} \label{app:implementation}

In this section, we show more implementation details on how we reduce the offload bound and save memory on both host and devices.

\paragraph{Continuous Host Buffer Bins} We notice that pytorch round up the size of host memory to nearest power of 2, resulting in at most 1x waste of host memory. To mitigate this for each offloading pass, we use bigger continuous buffer bins with sizes which are power of 2. We run a heuristic-based search to find a solution that \textbf{a)} All activation memory of a single offloading pass can fit these bins \textbf{b)} At most 3 bins are used \textbf{c)} The total size of the bins is minimum. In practice we find this method reduces the waste of host memory to a negligible amount. During reloading we move the continuous buffers directly to device and construct individual activation memory tensors on them.

\paragraph{Deterministic Device Memory Management} We allocate all offload-related buffers on the compute stream and use CUDA events to synchronize the usage of buffers between the offload stream and compute stream. This method circumvents the issue of non-deterministic buffer deallocation times that could prompt frequent cudaMallocs if one were to follow a more simplistic approach using Tensor.record\_stream.

\paragraph{Selective Recomputation with Negligible Overhead} \label{sec:selective_recomputation}

\begin{figure*}
    \centering
    \includegraphics[width=0.98\linewidth]{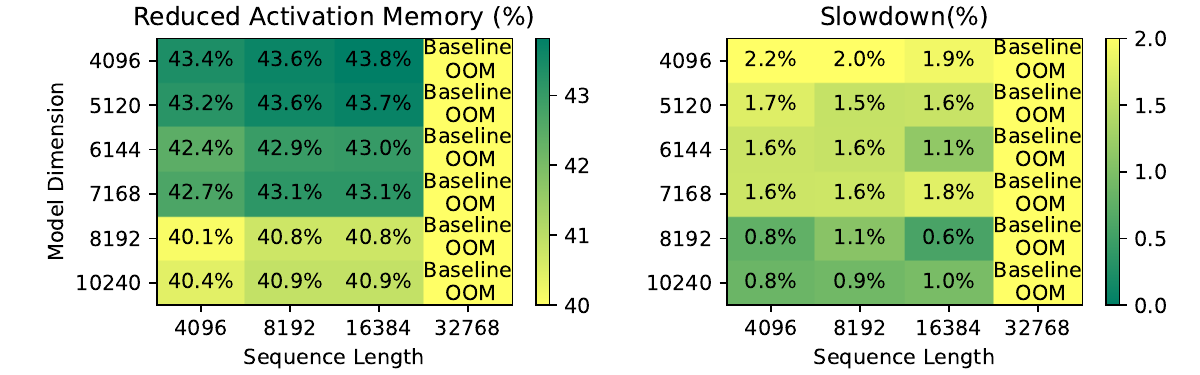}
    \caption{Ablation study on applying recomputation for Layernorm, GeLU and dropout layers. Approximately 40\% activation is reduced with around 1-2\% slowdown.}
    \label{fig:recompute_a100}
\end{figure*}

Using methods similar to \cite{chen2016training}, we simply recompute LayerNorm and GeLU layers in the backward pass to reduce the activation memory that is subject to offload. We also implement a customized dropout that preserves the random seed during the forward pass and reconstructs the dropout mask in the backward pass. Recomputing these layers reduces the activation memory for single layer from $34bsh$ (as \cite{korthikanti2023reducing}) to $20bsh$. Our ablation study results demonstrate a 40\% reduction in activation memory with negligible throughput overhead (around 1-2\% performance impact). We conduct an ablation experiment as in Figure \ref{fig:recompute_a100}

\paragraph{Topology-aware Offload Scheduling}
\label{sec:sync_interleaved_memory_transfer}
To achieve optimal performance, we implement a synchronized interleaved memory transfer schedule where two devices within the same NUMA node collaboratively execute alternating H2D and D2H transfers using cross-device event synchronization. The scheduling of memory transfer will be studied in Appendix \ref{sec:memory_transfer_ablation_study}. Our approach carefully co-designs computation and memory transfer schedules to ensure transfer stability and efficiency while maintaining interdependencies.

\paragraph{Enhanced Node Assignment} In many parallel processing (PP) schedules, there exists an imbalance in activation memory across stages, leading to a linear decrease in activation memory from the initial to the final rank. This imbalance places varying loads on the host memory capacity of different nodes. To optimize host memory utilization, we consolidate the PP stages by grouping lower-ranked stages with higher-ranked stages on the same node. For instance, in a PP setup with 16 devices, ranks 0-3 and 12-16 are assigned to node 0, while ranks 4-11 are allocated to node 1. It is important to note that this approach results in a doubling of cross-node communications.
\section{Topology-aware Offload Scheduling} \label{sec:memory_transfer_ablation_study}

\begin{figure*}[h!]
    \centering
    \includegraphics[width=0.98\linewidth]{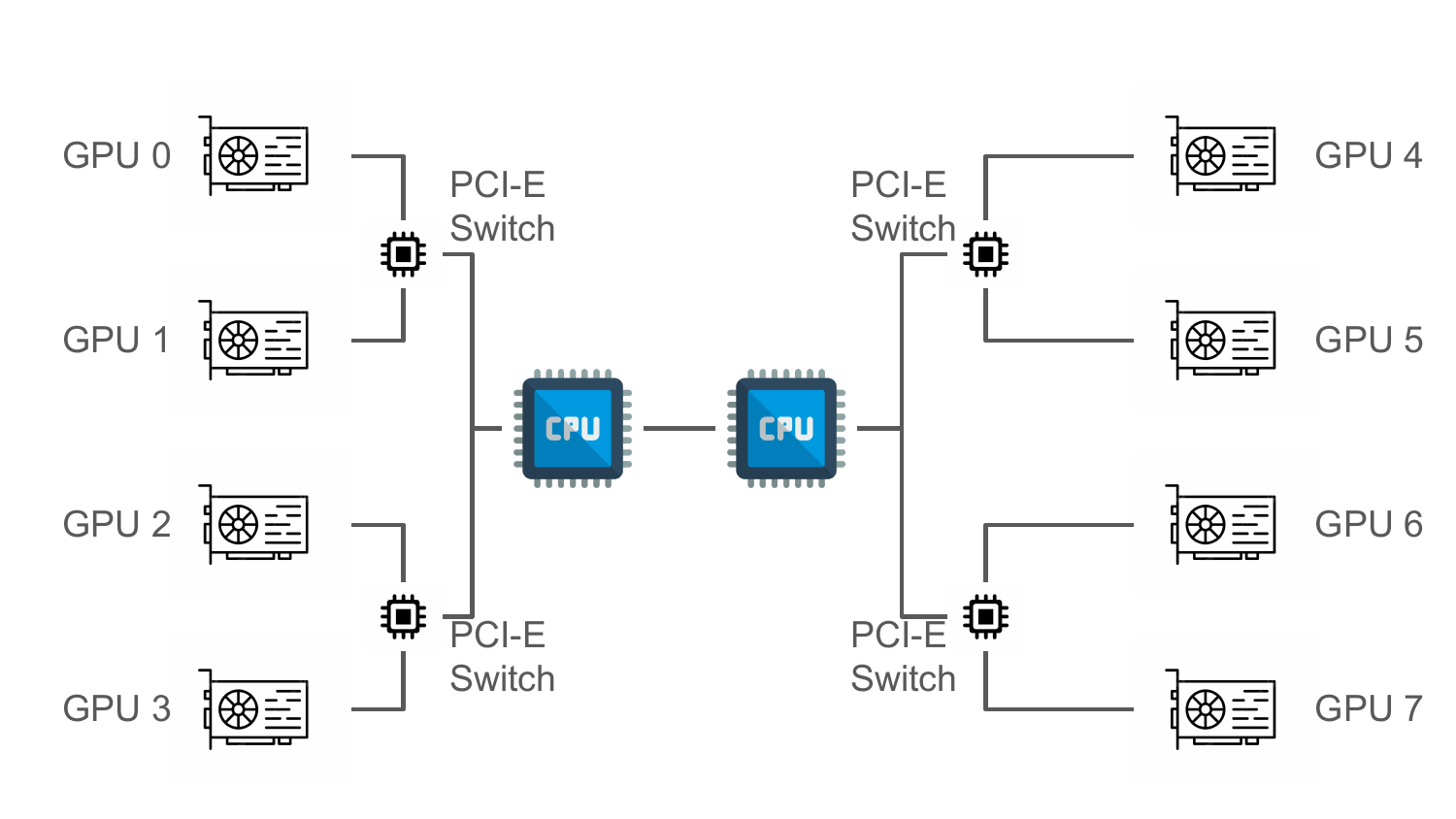}
    \caption{GPU server topology of our A100 GPU server. Adjcent GPUs share the same PCI-E switch hence potentially interfere with each other.}
    \label{fig:hardware_topology}
\end{figure*}
Designing the offload schedule based on the GPU server's topology is crucial. For instance, in our testing of the A100 server, illustrated in Figure \ref{fig:hardware_topology}, we observed that Host-to-Device (H2D) or Device-to-Host (D2H) transfers between GPU pairs on the same PCI-E switch could potentially interfere with each other. Therefore, coordinating the offload operations among GPU pairs on the same PCI-E switches becomes a critical consideration.

In our investigation, we assessed the throughput and stability of these transfers using various co-scheduling methods. We organized H2D and D2H operations into sequential groups, with operations within each group executing consecutively. To mimic real-world scenarios where two devices sharing the same PCI-E switch might trigger transfers at different times during training iterations, we introduced random inter-group delays. Each experiment was repeated multiple times, and we generated the distribution of per-device bidirectional throughput in Table \ref{table:offload_schedules}.

It's worth noting that while our experimental results favor the synchronized interleaved method, server topologies can differ. Therefore, in practice, it's essential to select co-scheduling methods that are most compatible with a specific hardware configuration.

% \subsection{Experimental Setup} \label{sec:memory_transfer_ablation_experiment_setup}

% Each training iteration consists of three distinct phases: warm-up, steady-state, and cool-down. The warm-up phase exclusively involves D2H transfers, while the cool-down phase handles H2D transfers. The steady-state phase encompasses both transfer types simultaneously. To determine the optimal schedule for each phase, we examine three scenarios: isolated D2H transfers, isolated H2D transfers, and concurrent H2D/D2H operations.

 %This experimental setup measures three key metrics: individual D2H transfer throughput, individual H2D transfer throughput, and total group throughput. We run them in groups so that the doubled value of the group throughput provides an approximation of the overall PCIe bidirectional throughput per NUMA node.

% In our settings, each H2D and D2H transfers 1 GB of data, and each group has 64 H2D or D2H operations. We ran our experiments on HPE ProLiant XL675d Gen10 Plus with 8× NVIDIA A100 80GB, AMD EPYC 7643 48-Core Processor, and PCIE Gen 4.

% \begin{figure*}
%     \centering
%     \includegraphics[width=0.98\linewidth]{figs/pcie_ablation.pdf}
%     \caption{Offload schedules tested in ablation study.}
%     \label{fig:pcie_ablation}
% \end{figure*}
\newcommand\cincludegraphics[2][]{\raisebox{-1\height}{\includegraphics[#1]{#2}}}
\begin{table}[h]
\caption{Comparing co-scheduling methods for offload operations on a pair of GPUs connected to the same PCI-E switch. Results shows that the syncronized interleaved schedule is both stable and fast.}
\label{table:offload_schedules}
\centering
\begin{tabular}{p{4cm}ScSc}
\hline
\textbf{Schedule} & \textbf{Sketch} & \textbf{Bandwidth Histogram} \\
\hline
\hline
\textbf{Parallel} Both devices under a NUMA node execute alternating H2D and D2H transfers parallelly. & \cincludegraphics[width=4cm]{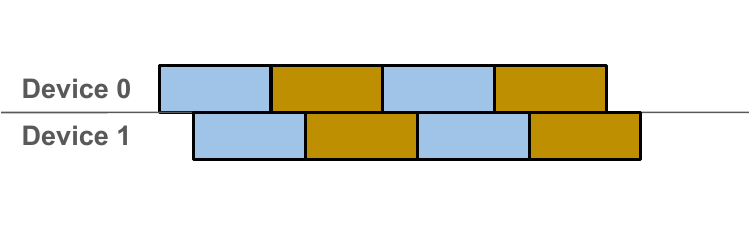} &  \cincludegraphics[width=4cm]{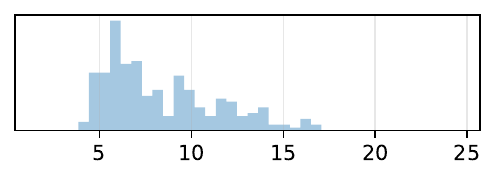}  \\
\hline
\textbf{Interleaved} Similar to parallel H2D-D2H, but devices under each NUMA node begin with complementary operations. & \cincludegraphics[width=4cm]{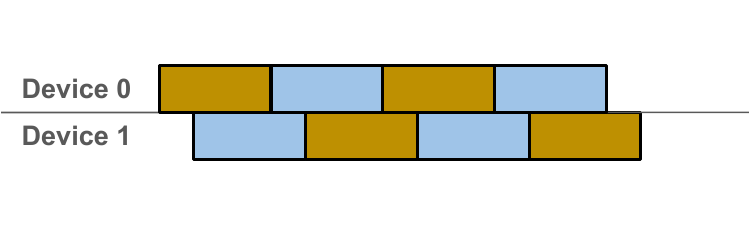}& \cincludegraphics[width=4cm]{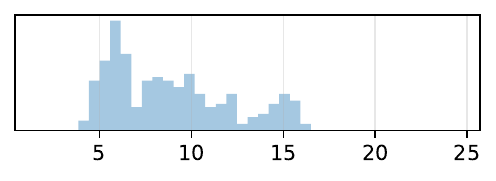}  \\
\hline
\textbf{Dual Stream} Both devices simultaneously execute H2D and D2H operations using separate CUDA streams. & \cincludegraphics[width=4cm]{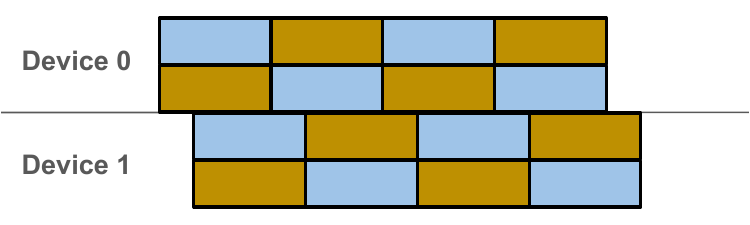}& \cincludegraphics[width=4cm]{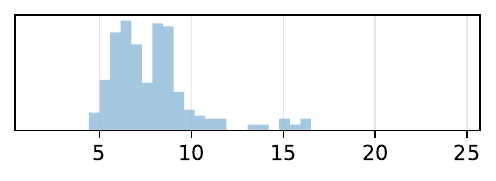} \\
\hline
\textbf{Synchronized Parallel} Parallel H2D-D2H with cross-device event synchronization before each memory transfer. & \cincludegraphics[width=4cm]{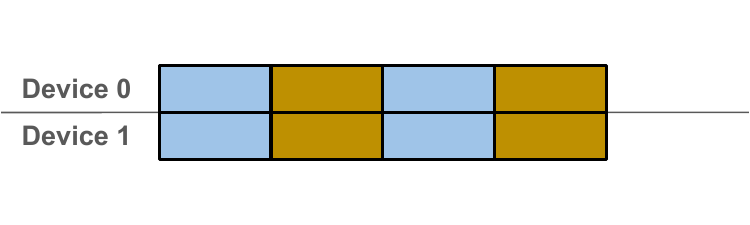} & \cincludegraphics[width=4cm]{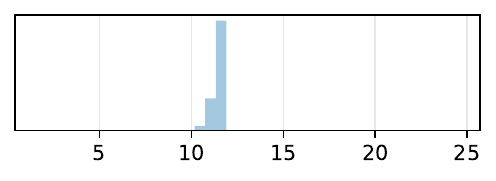}  \\
\hline
\textbf{Synchronized Interleaved} Interleaved H2D-D2H with cross-device event synchronization before each memory transfer. & \cincludegraphics[width=4cm]{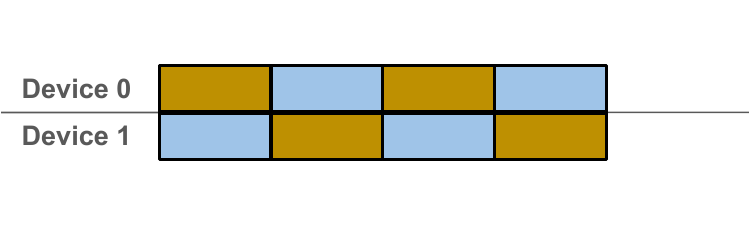}& \cincludegraphics[width=4cm]{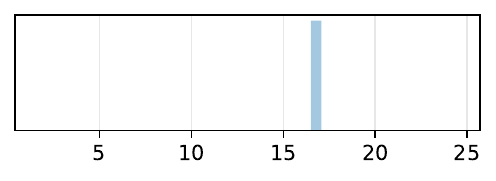}  \\
\hline
\textbf{Synchronized Dual Stream} Dual-Stream H2D-D2H with cross-device event synchronization before each memory transfer. & \cincludegraphics[width=4cm]{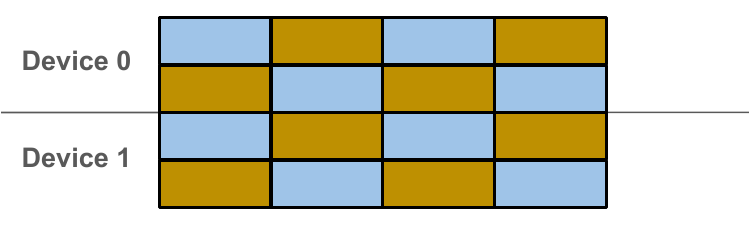}& \cincludegraphics[width=4cm]{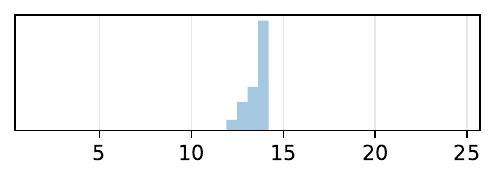}  \\
\hline

\end{tabular}
\end{table}

\section{Detailed Data on Activation Memory Scaling Study}
\begin{figure*}[h!]
    \centering
    \includegraphics[width=0.98\linewidth]{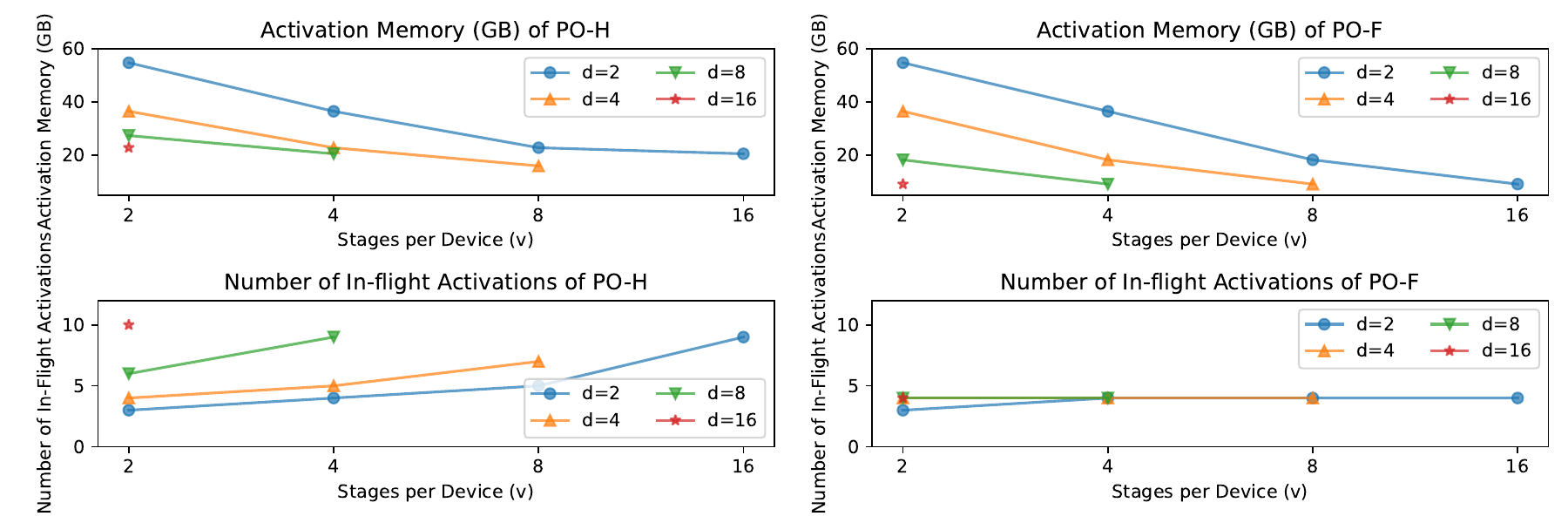}
    \caption{Scaling of activation memory under different $v$ and $d$.}
    \label{fig:scaling_detail}
\end{figure*}

In Section \ref{sec:mem_scaling} and Figure \ref{fig:scaling_best}, we've shown how activation memory scales with $v \times d$. In this section we show more detailed data on how the activation memory scale with $v$ and $d$ along, shown in a plot in Figure \ref{fig:scaling_detail}

%% file: example_paper.bbl
\begin{thebibliography}{23}
\providecommand{\natexlab}[1]{#1}
\providecommand{\url}[1]{\texttt{#1}}
\expandafter\ifx\csname urlstyle\endcsname\relax
  \providecommand{\doi}[1]{doi: #1}\else
  \providecommand{\doi}{doi: \begingroup \urlstyle{rm}\Url}\fi

\bibitem[Ainslie et~al.(2023)Ainslie, Lee-Thorp, de~Jong, Zemlyanskiy, Lebr{\'o}n, and Sanghai]{ainslie2023gqa}
Ainslie, J., Lee-Thorp, J., de~Jong, M., Zemlyanskiy, Y., Lebr{\'o}n, F., and Sanghai, S.
\newblock Gqa: Training generalized multi-query transformer models from multi-head checkpoints.
\newblock \emph{arXiv preprint arXiv:2305.13245}, 2023.

\bibitem[Chen et~al.(2016)Chen, Xu, Zhang, and Guestrin]{chen2016training}
Chen, T., Xu, B., Zhang, C., and Guestrin, C.
\newblock Training deep nets with sublinear memory cost.
\newblock \emph{arXiv preprint arXiv:1604.06174}, 2016.

\bibitem[Dubey et~al.(2024)Dubey, Jauhri, Pandey, Kadian, Al-Dahle, Letman, Mathur, Schelten, Yang, Fan, et~al.]{dubey2024llama}
Dubey, A., Jauhri, A., Pandey, A., Kadian, A., Al-Dahle, A., Letman, A., Mathur, A., Schelten, A., Yang, A., Fan, A., et~al.
\newblock The llama 3 herd of models.
\newblock \emph{arXiv preprint arXiv:2407.21783}, 2024.

\bibitem[Fan et~al.(2021)Fan, Rong, Meng, Cao, Wang, Zheng, Wu, Long, Yang, Xia, et~al.]{fan2021dapple}
Fan, S., Rong, Y., Meng, C., Cao, Z., Wang, S., Zheng, Z., Wu, C., Long, G., Yang, J., Xia, L., et~al.
\newblock Dapple: A pipelined data parallel approach for training large models.
\newblock In \emph{Proceedings of the 26th ACM SIGPLAN Symposium on Principles and Practice of Parallel Programming}, pp.\  431--445, 2021.

\bibitem[Goyal et~al.(2017)Goyal, Doll{\'a}r, Girshick, Noordhuis, Wesolowski, Kyrola, Tulloch, Jia, and He]{goyal2017accurate}
Goyal, P., Doll{\'a}r, P., Girshick, R., Noordhuis, P., Wesolowski, L., Kyrola, A., Tulloch, A., Jia, Y., and He, K.
\newblock Accurate, large minibatch sgd: Training imagenet in 1 hour.
\newblock \emph{arXiv preprint arXiv:1706.02677}, 2017.

\bibitem[Harlap et~al.(2018)Harlap, Narayanan, Phanishayee, Seshadri, Devanur, Ganger, and Gibbons]{harlap2018pipedream}
Harlap, A., Narayanan, D., Phanishayee, A., Seshadri, V., Devanur, N., Ganger, G., and Gibbons, P.
\newblock Pipedream: Fast and efficient pipeline parallel dnn training.
\newblock \emph{arXiv preprint arXiv:1806.03377}, 2018.

\bibitem[Huang et~al.(2019)Huang, Cheng, Bapna, Firat, Chen, Chen, Lee, Ngiam, Le, Wu, et~al.]{huang2019gpipe}
Huang, Y., Cheng, Y., Bapna, A., Firat, O., Chen, D., Chen, M., Lee, H., Ngiam, J., Le, Q.~V., Wu, Y., et~al.
\newblock Gpipe: Efficient training of giant neural networks using pipeline parallelism.
\newblock \emph{Advances in neural information processing systems}, 32, 2019.

\bibitem[Kim et~al.(2023)Kim, Kim, Yu, and Chun]{kim2023bpipe}
Kim, T., Kim, H., Yu, G.-I., and Chun, B.-G.
\newblock Bpipe: Memory-balanced pipeline parallelism for training large language models.
\newblock In \emph{International Conference on Machine Learning}, pp.\  16639--16653. PMLR, 2023.

\bibitem[Korthikanti et~al.(2023)Korthikanti, Casper, Lym, McAfee, Andersch, Shoeybi, and Catanzaro]{korthikanti2023reducing}
Korthikanti, V.~A., Casper, J., Lym, S., McAfee, L., Andersch, M., Shoeybi, M., and Catanzaro, B.
\newblock Reducing activation recomputation in large transformer models.
\newblock \emph{Proceedings of Machine Learning and Systems}, 5, 2023.

\bibitem[Li et~al.(2021)Li, Zhuang, Guo, Zhuo, Zhang, Song, and Stoica]{li2021terapipe}
Li, Z., Zhuang, S., Guo, S., Zhuo, D., Zhang, H., Song, D., and Stoica, I.
\newblock Terapipe: Token-level pipeline parallelism for training large-scale language models.
\newblock In \emph{International Conference on Machine Learning}, pp.\  6543--6552. PMLR, 2021.

\bibitem[Liu et~al.(2024)Liu, Feng, Xue, Wang, Wu, Lu, Zhao, Deng, Zhang, Ruan, et~al.]{liu2024deepseek}
Liu, A., Feng, B., Xue, B., Wang, B., Wu, B., Lu, C., Zhao, C., Deng, C., Zhang, C., Ruan, C., et~al.
\newblock Deepseek-v3 technical report.
\newblock \emph{arXiv preprint arXiv:2412.19437}, 2024.

\bibitem[Liu et~al.(2023)Liu, Cheng, Zhou, and You]{Liu2023HanayoHW}
Liu, Z., Cheng, S., Zhou, H., and You, Y.
\newblock Hanayo: Harnessing wave-like pipeline parallelism for enhanced large model training efficiency.
\newblock \emph{The International Conference for High Performance Computing, Networking, Storage, and Analysis}, pp.\  1--13, 2023.
\newblock URL \url{https://api.semanticscholar.org/CorpusID:261339639}.

\bibitem[Narayanan et~al.(2021)Narayanan, Shoeybi, Casper, LeGresley, Patwary, Korthikanti, Vainbrand, Kashinkunti, Bernauer, Catanzaro, et~al.]{narayanan2021efficient}
Narayanan, D., Shoeybi, M., Casper, J., LeGresley, P., Patwary, M., Korthikanti, V., Vainbrand, D., Kashinkunti, P., Bernauer, J., Catanzaro, B., et~al.
\newblock Efficient large-scale language model training on gpu clusters using megatron-lm.
\newblock In \emph{Proceedings of the International Conference for High Performance Computing, Networking, Storage and Analysis}, pp.\  1--15, 2021.

\bibitem[Qi et~al.(2023)Qi, Wan, Huang, and Lin]{qi2023zero}
Qi, P., Wan, X., Huang, G., and Lin, M.
\newblock Zero bubble pipeline parallelism.
\newblock In \emph{The Twelfth International Conference on Learning Representations}, 2023.

\bibitem[Qi et~al.(2024)Qi, Wan, Amar, and Lin]{qi2024pipeline}
Qi, P., Wan, X., Amar, N., and Lin, M.
\newblock Pipeline parallelism with controllable memory.
\newblock \emph{arXiv preprint arXiv:2405.15362}, 2024.

\bibitem[Rajbhandari et~al.(2020)Rajbhandari, Rasley, Ruwase, and He]{rajbhandari2020zero}
Rajbhandari, S., Rasley, J., Ruwase, O., and He, Y.
\newblock Zero: Memory optimizations toward training trillion parameter models.
\newblock In \emph{SC20: International Conference for High Performance Computing, Networking, Storage and Analysis}, pp.\  1--16. IEEE, 2020.

\bibitem[Rajbhandari et~al.(2021)Rajbhandari, Ruwase, Rasley, Smith, and He]{rajbhandari2021zero}
Rajbhandari, S., Ruwase, O., Rasley, J., Smith, S., and He, Y.
\newblock Zero-infinity: Breaking the gpu memory wall for extreme scale deep learning.
\newblock In \emph{Proceedings of the international conference for high performance computing, networking, storage and analysis}, pp.\  1--14, 2021.

\bibitem[Ren et~al.(2021)Ren, Rajbhandari, Aminabadi, Ruwase, Yang, Zhang, Li, and He]{ren2021zero}
Ren, J., Rajbhandari, S., Aminabadi, R.~Y., Ruwase, O., Yang, S., Zhang, M., Li, D., and He, Y.
\newblock $\{$Zero-offload$\}$: Democratizing $\{$billion-scale$\}$ model training.
\newblock In \emph{2021 USENIX Annual Technical Conference (USENIX ATC 21)}, pp.\  551--564, 2021.

\bibitem[Shoeybi et~al.(2019)Shoeybi, Patwary, Puri, LeGresley, Casper, and Catanzaro]{shoeybi2019megatron}
Shoeybi, M., Patwary, M., Puri, R., LeGresley, P., Casper, J., and Catanzaro, B.
\newblock Megatron-lm: Training multi-billion parameter language models using model parallelism.
\newblock \emph{arXiv preprint arXiv:1909.08053}, 2019.

\bibitem[Sun et~al.(2024)Sun, Zhao, Han, Yang, Zhang, Liu, Shi, and Sun]{sun2024seq1f1b}
Sun, A., Zhao, W., Han, X., Yang, C., Zhang, X., Liu, Z., Shi, C., and Sun, M.
\newblock Seq1f1b: Efficient sequence-level pipeline parallelism for large language model training.
\newblock \emph{arXiv preprint arXiv:2406.03488}, 2024.

\bibitem[Vaswani et~al.(2017)Vaswani, Shazeer, Parmar, Uszkoreit, Jones, Gomez, Kaiser, and Polosukhin]{vaswani2017attention}
Vaswani, A., Shazeer, N., Parmar, N., Uszkoreit, J., Jones, L., Gomez, A.~N., Kaiser, {\L}., and Polosukhin, I.
\newblock Attention is all you need.
\newblock \emph{Advances in neural information processing systems}, 30, 2017.

\bibitem[Yeung et~al.(2024)Yeung, Qi, Lin, and Wan]{yeung2024balancing}
Yeung, M.~T., Qi, P., Lin, M., and Wan, X.
\newblock Balancing pipeline parallelism with vocabulary parallelism.
\newblock \emph{arXiv preprint arXiv:2411.05288}, 2024.

\bibitem[Yuan et~al.(2024)Yuan, Liu, Ye, Zhang, Tan, Chen, Song, and Zhang]{yuan2024accelerating}
Yuan, T., Liu, Y., Ye, X., Zhang, S., Tan, J., Chen, B., Song, C., and Zhang, D.
\newblock Accelerating the training of large language models using efficient activation rematerialization and optimal hybrid parallelism.
\newblock In \emph{2024 USENIX Annual Technical Conference (USENIX ATC 24)}, pp.\  545--561, 2024.

\end{thebibliography}
